\documentclass[conference]{IEEEtran}
\usepackage{graphicx}
\usepackage{subfigure}
\usepackage{standalone}
\usepackage[textsize=tiny]{todonotes}
\usepackage{tabularx}
\usepackage{pgfplots, pgfplotstable}
\usepackage{textcomp}
\usepackage{makecell}
\usepackage{booktabs}
\usepackage{listings}
\usepackage{lipsum}
\usepackage{multirow}
\usepackage{color}
\usepackage{cite}
\usepackage{amsmath,amssymb,amsfonts}
\usepackage{algorithmic}
\usepackage{xcolor}
\usepackage{multirow}
\usepackage{subfigure}
\usepackage{diagbox}
\usepackage[T1]{fontenc}
\makeatletter
\newcommand\BeraMonottfamily{%
  \def\fvm@Scale{0.85}
  \fontfamily{fvm}\selectfont
}
\makeatother

\usepackage{xcolor}
\usepackage{array}
\usepackage{wasysym}
\usepackage[flushleft]{threeparttable}

\begin{document}

\title{Perceptual Quality Assessment for Fine-Grained Compressed Images}

\author{
 \IEEEauthorblockN{Zicheng Zhang, Wei Sun,  Wei Wu, Ying Chen, \emph{Senior Member, IEEE}, \\ Xiongkuo Min, \emph{Member, IEEE} and Guangtao Zhai,  \emph{Senior Member, IEEE}}
}

\maketitle

\thispagestyle{plain}
\pagestyle{plain}

\begin{abstract}
Recent years have witnessed the rapid development of image storage and transmission systems, in which image compression plays an important role. Generally speaking, image compression algorithms are developed to ensure good visual quality at limited bit rates. However, due to the different compression optimization methods, the compressed images may have different levels of quality, which needs to be evaluated quantificationally. Nowadays, the mainstream full-reference (FR) metrics are effective to predict the quality of compressed images at coarse-grained levels (the bit rates differences of compressed images are obvious), however, they may perform poorly for fine-grained compressed images whose bit rates differences are quite subtle. Therefore, to better improve the Quality of Experience (QoE) and provide useful guidance for compression algorithms, we propose a full-reference image quality assessment (FR-IQA) method for compressed images of fine-grained levels. Specifically, the reference images and compressed images are first converted to $YCbCr$ color space. The gradient features are extracted from regions that are sensitive to compression artifacts. Then we employ the Log-Gabor transformation to further analyze the texture difference. Finally, the obtained features are fused into a quality score. The proposed method is validated on the fine-grained compression image quality assessment (FGIQA) database, which is especially constructed for assessing the quality of compressed images with close bit rates. The experimental results show that our metric outperforms mainstream FR-IQA metrics on the FGIQA database. We also test our method on other commonly used compression IQA databases and the results show that our method obtains competitive performance on the coarse-grained compression IQA databases as well.

\end{abstract}


\let\thefootnote\relax\footnotetext{

Zicheng Zhang, Wei Sun, Xiongkuo Min, and Guangtao Zhai are with the Institute of Image Communication and Network Engineering, Shanghai Jiao Tong University, 200240 Shanghai, China. E-mail:\{zzc1998,sunguwei,minxiongkuo,zhaiguangtao\}@sjtu.edu.cn 

Wei Wu and Ying Chen are with the Alibaba Group, 310052 Hangzhou, China. 
E-mail: \{guokui.ww, chenying.ailab\}@alibaba-inc.com.

}

\bstctlcite{IEEEexample:BSTcontrol}

\IEEEpeerreviewmaketitle

\section{Introduction}
\subsection{Motivation}
With the rapid development of mobile devices and social media, the last decade has witnessed the blowout development of image services. To improve the viewer's Quality of Experience (QoE), images are expected to be transmitted with better-perceived quality under the constraints of limited bandwidth and demand for lower latency, which brings a huge challenge for image compression techniques along with the quality assessment for image compression \cite{c2,c5,c3,c4,c1,c6,jpeg1}. Although the visual quality of compressed images is extremely sensitive to the bit rate (higher bit rate usually indicates higher visual quality), the bit rate is fixed for image compression in most situations, which means the image compression systems need to obtain better visual quality of compressed images with fixed bit-rate level. In this paper, coarse-grained compressed images refer to images that are compressed with obviously different bit rates while fine-grained compressed images represent images that are compressed with close or the same bit rate.

\begin{figure}[t]
    \centering
    \subfigure[Coarse-grained1]{
    \begin{minipage}[t]{0.45\linewidth}
    \centering
    \includegraphics[width=3.8cm]{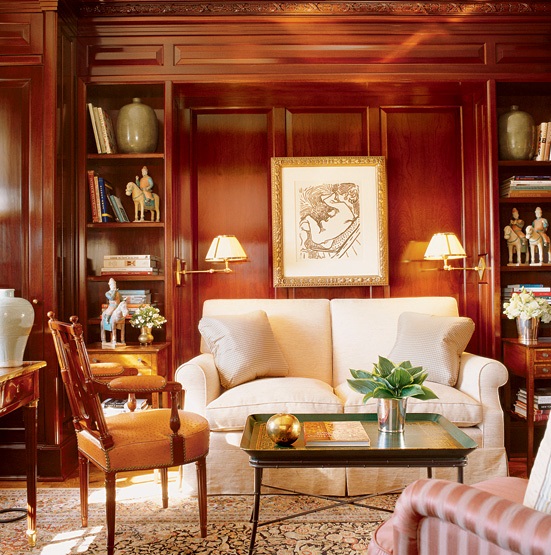}
    \end{minipage}%
    }%
    \subfigure[Coarse-grained2]{
    \begin{minipage}[t]{0.45\linewidth}
    \centering
    \includegraphics[width=3.8cm]{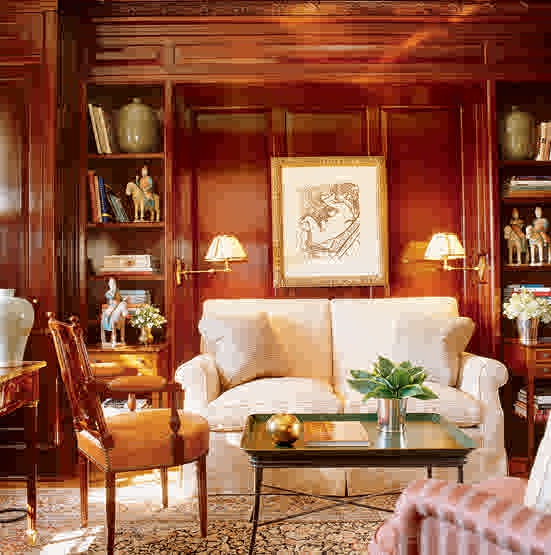}
    \end{minipage}%
    }%
    
    \subfigure[Fine-grained1]{
    \begin{minipage}[t]{0.45\linewidth}
    \centering
    \includegraphics[width=3.8cm]{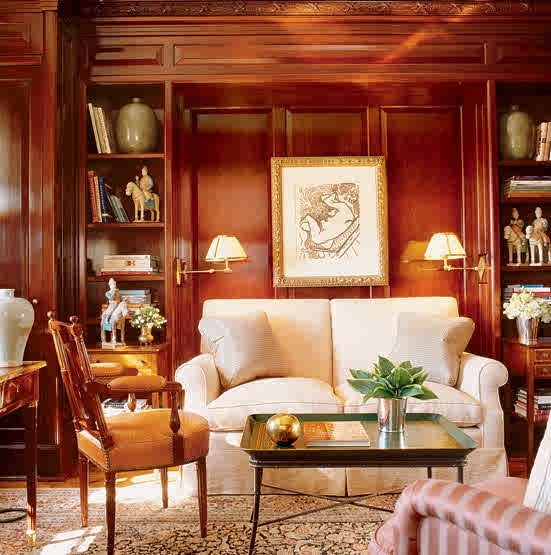}
    \end{minipage}%
    }%
    \subfigure[Fine-grained2]{
    \begin{minipage}[t]{0.45\linewidth}
    \centering
    \includegraphics[width=3.8cm]{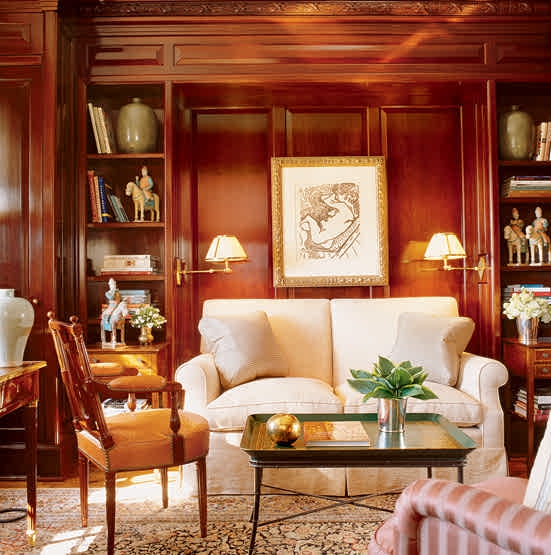}
    \end{minipage}%
    }%
    \caption{Example of images with fine-grained and coarse-grained compression levels. Specifically, (a) and (b) represent the images with coarse-grained compression levels, and (c) and (d) are images compressed with fine-grained levels.}
    \label{fig:introduction}
    \vspace{-0.5cm}
\end{figure}

JPEG and JPEG2000 are two of the most highly accepted image compression techniques, which achieve required compression rates by introducing information loss to the compressed images \cite{jpeg}. In these image coding standards, various coding parameters and modes are utilized to compress the images, which can lead to distinct performance improvements at fixed bit rate. In JPEG coding standard \cite{jpeg2}, the customized quantization table is established based on subjective judgment. The open-source and more developed JPEG library, lib-jpeg \cite{lib-jpeg}, further employs 8 quantization tables based on various optimization criteria. In these compression systems, MSE (mean squared error) and PSNR (peak-signal-to-noise-ratio) are mostly utilized as the quality assessment metrics and optimization criterion \cite{mse1}\cite{mse2}. However, it has been proven that PSNR and MSE may not be effective enough to reflect the human vision system (HVS) on the quality of fine-grained compressed images \cite{fgiqa}, which indicates that the compression systems using such criterion may sacrifice the quality-aware information to compromise with the compression bit rate. Therefore, better quality assessment methods are needed to optimize such compression systems.

\vspace{0.3cm}

\subsection{Related Works}

Image quality assessment (IQA) is developed to evaluate the perceived visual quality based on human vision system (HVS) \cite{iqa1,iqa2,iqa3,iqa4,zhang2021no,gu2014using,gu2015analysis,fang2014no,gu2015no,zhai2011psychovisual,gu2014hybrid}, which has a wide range of application in many image processes including image compression. To keep more consistent with HVS, a large number of IQA metrics have been developed in recent decades \cite{ssim,vsi,ms-ssim,iw-ssim,fsim,mad,gsmd,ifc,bayesian,pieapp,lpips}. Generally speaking, IQA can be divided into subjective quality assessment and objective quality assessment according to the involvement of humans \cite{iqa4}. The subjective quality assessment is more accurate but is very expensive and time-consuming, therefore, objective quality assessment gains more research interest. Specifically, objective quality assessment can be divided into three categories: full-reference (FR) IQA methods, reduced-reference (RR) IQA methods, and no-reference (NR) IQA methods \cite{xiong1}. FR-IQA methods work only when the pristine reference image is available while RR-IQA and NR-IQA methods operate with part of reference information and completely no reference information. Despite that NR-IQA methods are more robust facing various application scenarios, the reference images are usually easy to access in compression systems. Thus in this paper, we mainly focus on FR-IQA methods, which are more commonly used in compression systems.

The popular peak-signal-to-noise-ratio (PSNR) index quantifies reconstruction quality for images by measuring the difference between pixels. Another well-known FR-IQA method is structural similarity index (SSIM) \cite{ssim}, which calculates the quality scores by comparing the correlation of luminance and contrast based on the assumption that HVS is sensitive to the local structures. Then some extended SSIM methods are proposed. Multi-Scale SSIM (MS-SSIM) \cite{ms-ssim} makes further improvements by calculating SSIM of different image scales. Feature similarity index (FSIM) \cite{fsim} combines phase and gradient features' similarity to obtain the quality levels of distorted images. Information Weighted SSIM (IW-SSIM) \cite{iw-ssim} uses spatially varied weights to gain better performance. Gradient Magnitude Similarity Deviation (GMSD) \cite{gsmd} achieves good performance by making use of gradient magnitude information. Apart from SSIM's extensions, Visual Saliency-Induced index (VSI) \cite{vsi} proposes to employ saliency information to optimize the quality assessment procedure. Information Fidelity Criterion (IFC) \cite{ifc} utilizes the natural scene statistics (NSS) to quantify the distortions of images, which proves to be very effective for IQA tasks. With the development huge success of convolution neural networks (CNN), some deep learning-based methods have been proposed to deal with IQA problems, such as PieAPP \cite{pieapp} and LPIPS \cite{lpips}. These methods achieve outstanding performance, however, they heavily rely on the large-scale IQA databases and may be less effective to predict the quality of images outside the range of training.

\begin{figure*}
    \centering
    \includegraphics[width = 17cm]{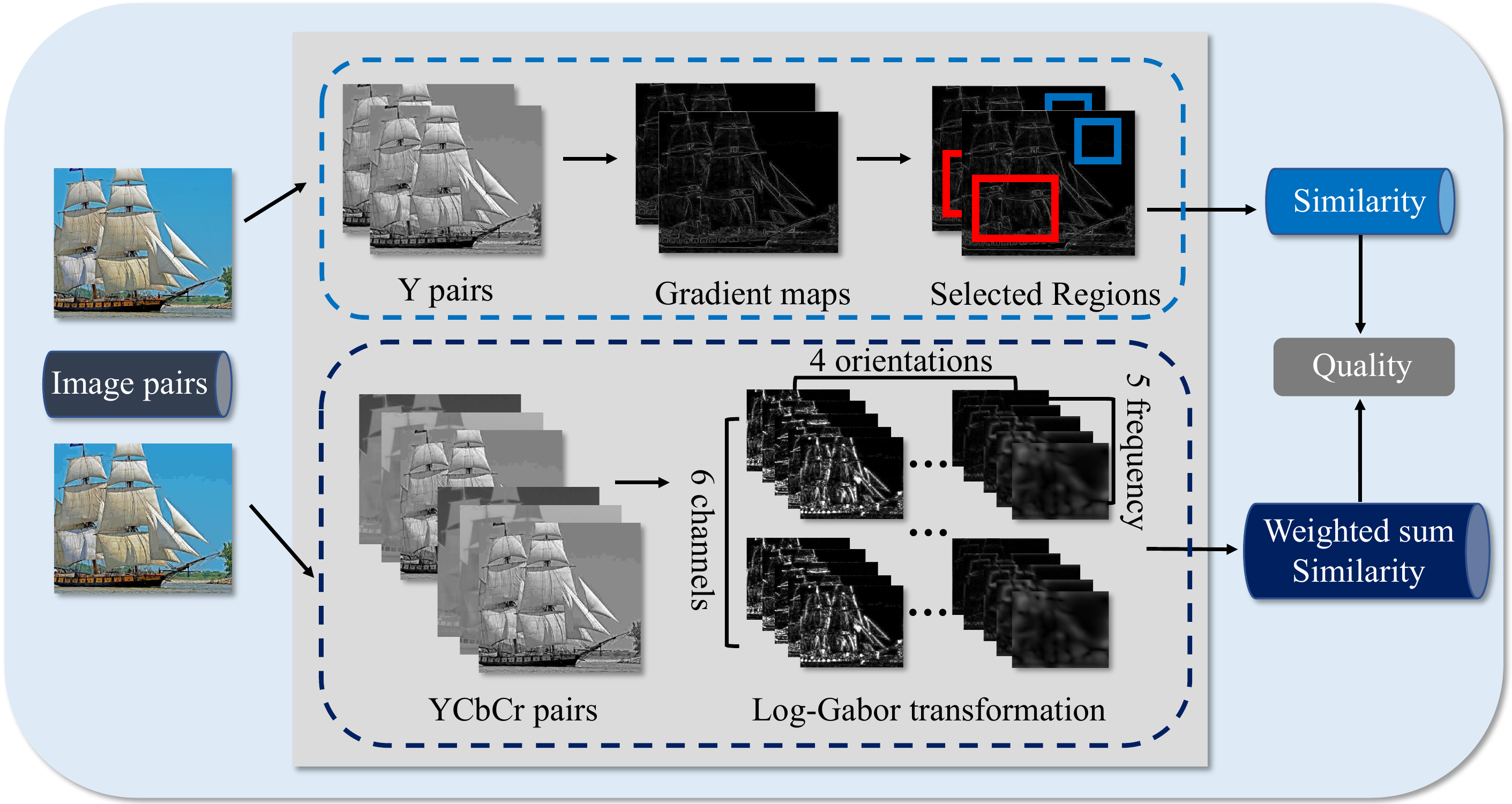}
    \caption{The framework of the proposed method.}
    \label{fig:framework}
\end{figure*}

The FR-IQA methods mentioned above are proven to be consistent with human perception and are able to accurately predict the quality levels of distorted images, including compressed images. Although they have obtained excellent performance on IQA databases like LIVE \cite{live}, TID2013 \cite{tid2013}, and CSIQ \cite{csiq}, they fail to correctly evaluate the quality of fine-grained compressed images with subtle differences in some piratical situations, which is demonstrated in Section \ref{sec:experiment}. The reason is that the images in such IQA databases are usually compressed at relatively coarse-grained levels, which means that the bit rate and compression distortions of evaluated images are obviously different. Therefore, the IQA methods developed on such coarse-grained databases may not be effective for fine-grained compressed images. This brings a great challenge for image compression quality assessment because the compression systems are often required to obtain the satisfying visual quality of images at fixed bit rate. As illustrated in Fig. \ref{fig:introduction}, the image (a) is compressed at a bit rate of 2.67 bits per pixel (bpp) and the image (b) is compressed at a bit rate of 1.96 bpp. It can be easily observed that these coarse-grained compressed images have obvious distortion differences such as blur and structural damage, and we can confidently draw the conclusion that image (a) obtains better visual quality than image (b). The images (c) and (d) are compressed at a bit rate of about 2.37 bpp using two different optimization methods and it is relatively difficult to distinguish the visual quality differences. 

\vspace{0.3cm}
\subsection{Our Approach}

Normally speaking, people usually perceive the compressed images in both coarse and fine manners. They not only assess the obvious compression artifacts such as blur and structure damage, at the same time, they also judge the quality through the indistinct details of compressed images.
Therefore, to correspond with the human perception process of compressed images, we propose a novel FR-IQA metric for compressed images by extracting coarse and fine features from two aspects: image gradient and texture. 

The gradient information is sensitive to obvious distortions such as structural damage and block effect \cite{gsmd,dehaze1} while texture information can help further identify the tiny compression artifacts such as texture shift \cite{texture1,texture2}. Considering such situations, the reference images and compressed images are first converted from $RGB$ color space to $YCbCr$ color space, which has been employed in many compression systems. Then the gradient-based features are extracted by calculating the statistical parameters of the gradient similarity maps in selected regions. Specifically, we employ frequency information obtained also by image gradient to describe the distortion-aware regions. The texture-based features are extracted by computing the statistical parameters of the Log-Gabor similarity maps with different frequency levels and orientations. Finally, the quality score is obtained by fusing the feature similarities mentioned above. 

The proposed method is tested on the fine-grained compression IQA (FGIQA) database and the experimental results show that our method achieves a better correlation with HVS than other state-of-the-art FR-IQA methods. The ablation experiment and statistical test further confirm the contributions of different feature groups and the effectiveness of our method. We also validate our method on other popular IQA databases in which the images are compressed at relatively coarse-grained levels and the results show that the proposed method obtains strong generalization ability on coarse-grained compressed images.

This paper is based on our previous paper \cite{zhang-vcip}. Compared with previous work, we make several new contributions. Most of the contributions have been described above, and a summarization is given here. First, we connect the process of extracting features with the coarse and fine perception manners. Second, we make more detailed discussions and show more figures in the proposed method explanation. Third, we conducted the parameter sensitivity experiment to obtain the optimal parameters. Fourth, the statistical test and ablation experiment are conduct to further test the effectiveness and contributions of the extracted features in the proposed method. 

The rest of the paper is organized as follows. Section \ref{sec:proposed} describes the proposed method. Section \ref{sec:experiment} presents the experiment setup and the experimental results. Section \ref{sec:conclusion} concludes the paper. 

\begin{figure*}[t]
    \centering
    \subfigure[The reference image.]{
    \begin{minipage}[t]{0.19\linewidth}
    \centering
    \includegraphics[width=3.4cm, height = 3cm]{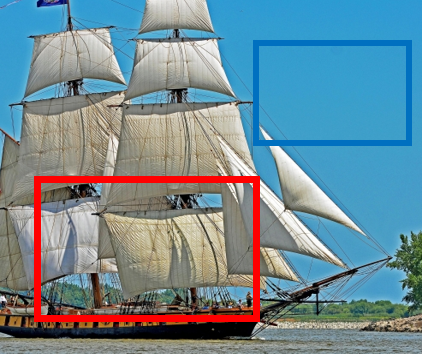}
    \end{minipage}%
    }%
    \subfigure[]{
    \begin{minipage}[t]{0.19\linewidth}
    \centering
    \includegraphics[width=3.4cm, height = 3cm]{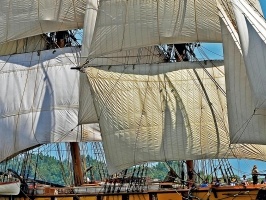}
    \end{minipage}%
    }%
    \subfigure[]{
    \begin{minipage}[t]{0.19\linewidth}
    \centering
    \includegraphics[width=3.4cm, height = 3cm]{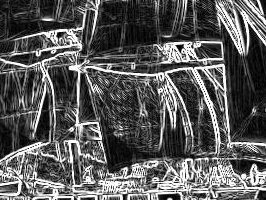}
    \end{minipage}%
    }%
    \subfigure[]{
    \begin{minipage}[t]{0.19\linewidth}
    \centering
    \includegraphics[width=3.4cm, height = 3cm]{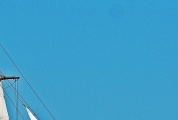}
    \end{minipage}%
    }%
    \subfigure[]{
    \begin{minipage}[t]{0.19\linewidth}
    \centering
    \includegraphics[width=3.4cm, height = 3cm]{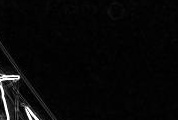}
    \end{minipage}%
    }%
    
    \subfigure[The distorted image.]{
    \begin{minipage}[t]{0.19\linewidth}
    \centering
    \includegraphics[width=3.4cm, height = 3cm]{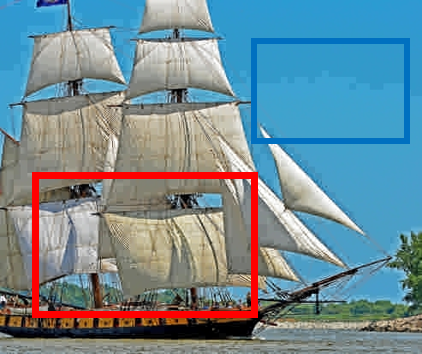}
    \end{minipage}%
    }%
    \subfigure[]{
    \begin{minipage}[t]{0.19\linewidth}
    \centering
    \includegraphics[width=3.4cm, height = 3cm]{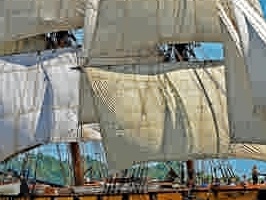}
    \end{minipage}%
    }%
    \subfigure[]{
    \begin{minipage}[t]{0.19\linewidth}
    \centering
    \includegraphics[width=3.4cm, height = 3cm]{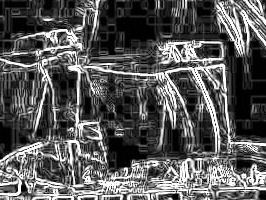}
    \end{minipage}%
    }%
    \subfigure[]{
    \begin{minipage}[t]{0.19\linewidth}
    \centering
    \includegraphics[width=3.4cm, height = 3cm]{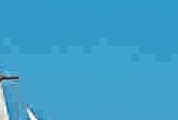}
    \end{minipage}%
    }%
    \subfigure[]{
    \begin{minipage}[t]{0.19\linewidth}
    \centering
    \includegraphics[width=3.4cm, height = 3cm]{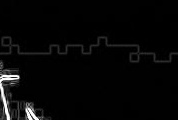}
    \end{minipage}%
    }%
    \caption{An example of regions with compression distortion. The red rectangle illustrates the high-frequency area where ringing and blur occur, while the blue rectangle illustrates the low-frequency area where block effect and structure over-restoration can be seen. (b) and (g) are enlarged illustrations of red rectangle areas while (d) and (i) are enlarged illustrations of blue areas respectively. (c), (e), (h), and (j) are the corresponding gradient maps for the above illustrations. }
    \label{fig:gradient}
\end{figure*}

\begin{figure*}[t]
    \centering
    \subfigure[Ref]{
    \begin{minipage}[t]{0.19\linewidth}
    \centering
    \includegraphics[width=3.4cm]{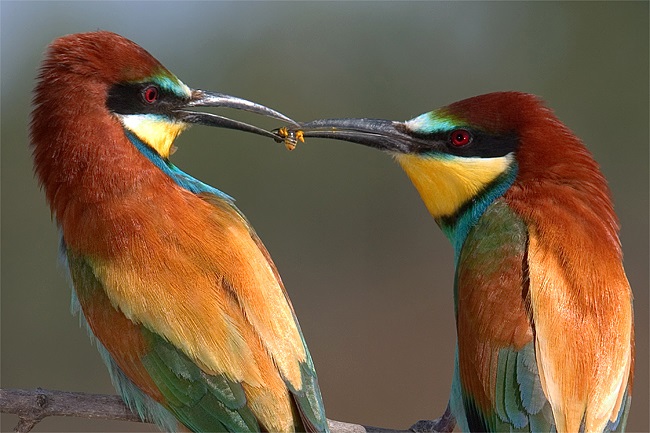}
    \end{minipage}%
    }%
    \subfigure[$0^{\circ}$]{
    \begin{minipage}[t]{0.19\linewidth}
    \centering
    \includegraphics[width=3.4cm]{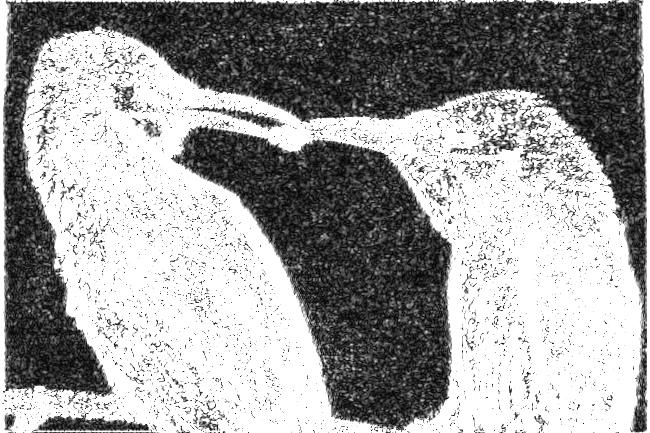}
    \end{minipage}%
    }%
    \subfigure[$45^{\circ}$]{
    \begin{minipage}[t]{0.19\linewidth}
    \centering
    \includegraphics[width=3.4cm]{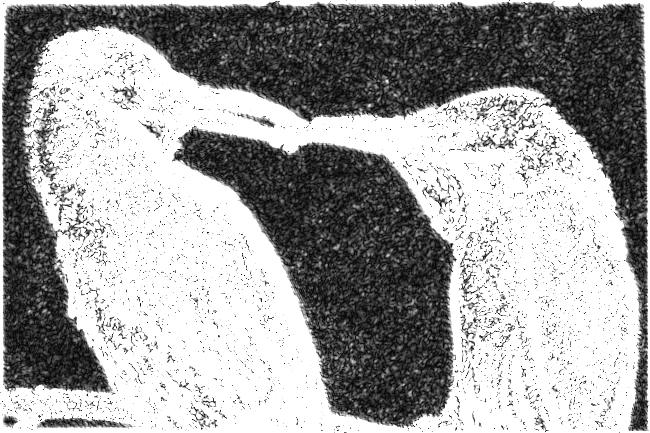}
    \end{minipage}%
    }%
    \subfigure[$90^{\circ}$]{
    \begin{minipage}[t]{0.19\linewidth}
    \centering
    \includegraphics[width=3.4cm]{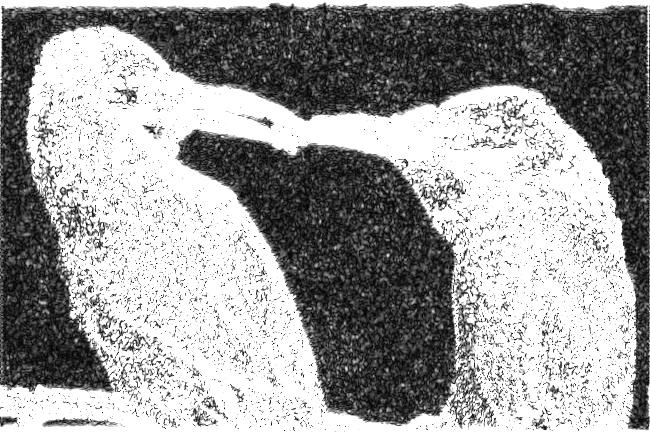}
    \end{minipage}%
    }%
    \subfigure[$135^{\circ}$]{
    \begin{minipage}[t]{0.19\linewidth}
    \centering
    \includegraphics[width=3.4cm]{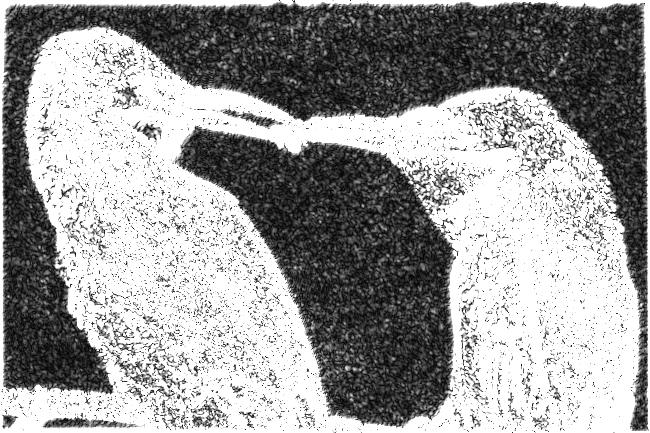}
    \end{minipage}%
    }%

    \subfigure[Dis1]{
    \begin{minipage}[t]{0.19\linewidth}
    \centering
    \includegraphics[width=3.4cm]{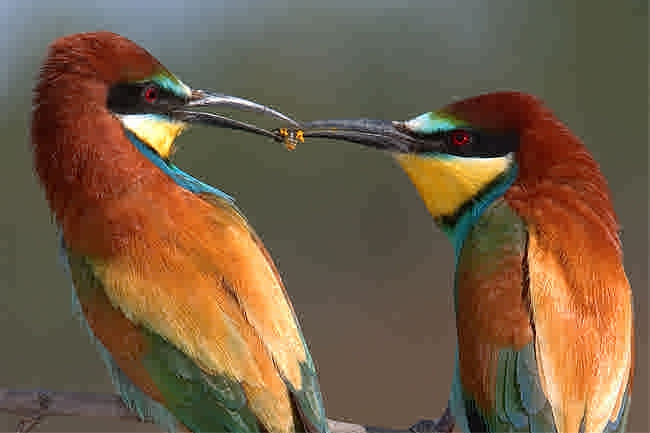}
    \end{minipage}%
    }%
    \subfigure[$0^{\circ}$]{
    \begin{minipage}[t]{0.19\linewidth}
    \centering
    \includegraphics[width=3.4cm]{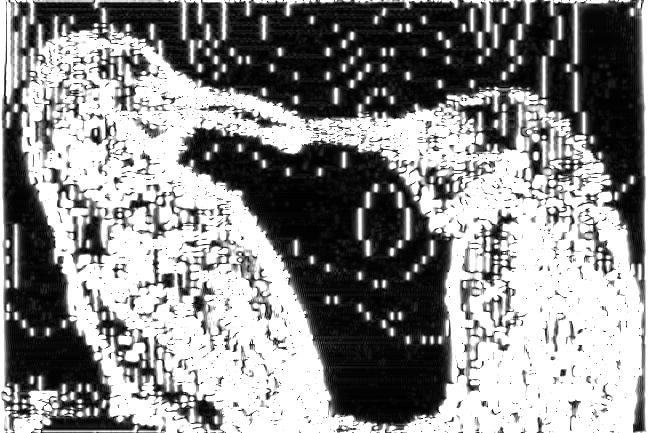}
    \end{minipage}%
    }%
    \subfigure[$45^{\circ}$]{
    \begin{minipage}[t]{0.19\linewidth}
    \centering
    \includegraphics[width=3.4cm]{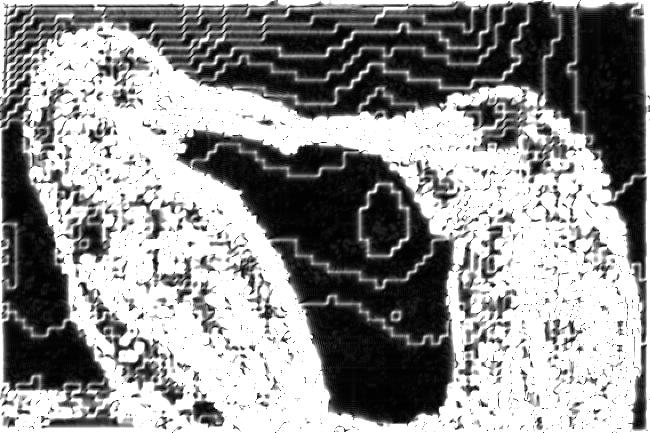}
    \end{minipage}%
    }%
    \subfigure[$90^{\circ}$]{
    \begin{minipage}[t]{0.19\linewidth}
    \centering
    \includegraphics[width=3.4cm]{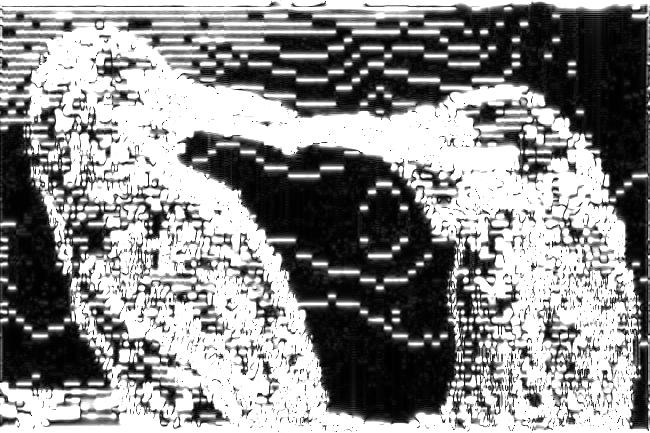}
    \end{minipage}%
    }%
    \subfigure[$135^{\circ}$]{
    \begin{minipage}[t]{0.19\linewidth}
    \centering
    \includegraphics[width=3.4cm]{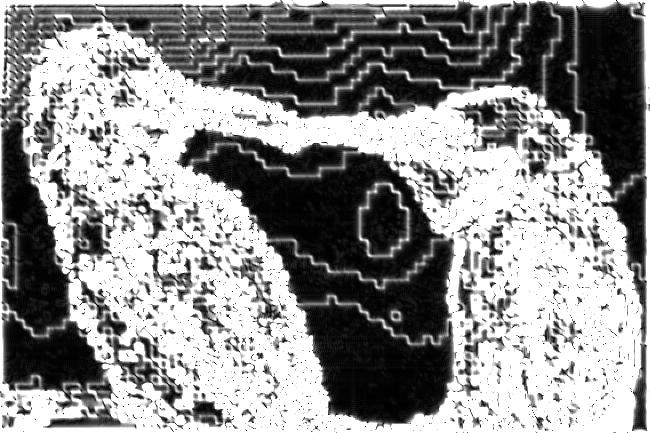}
    \end{minipage}
    }%
    
    \subfigure[Dis2]{
    \begin{minipage}[t]{0.19\linewidth}
    \centering
    \includegraphics[width=3.4cm]{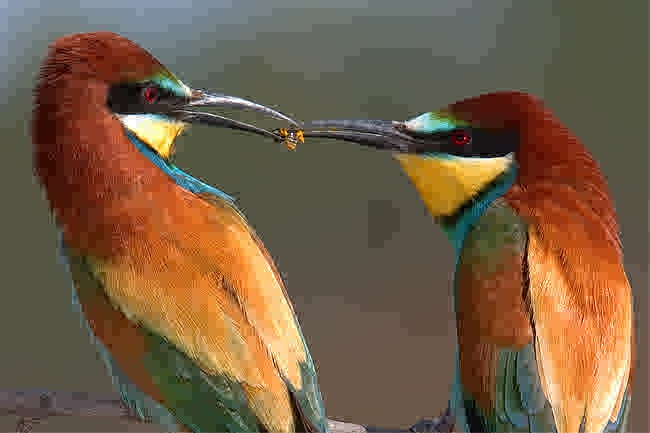}
    \end{minipage}%
    }%
    \subfigure[$0^{\circ}$]{
    \begin{minipage}[t]{0.19\linewidth}
    \centering
    \includegraphics[width=3.4cm]{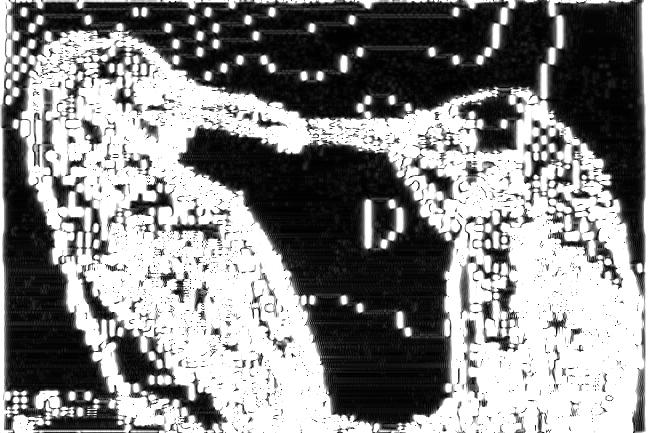}
    \end{minipage}%
    }%
    \subfigure[$45^{\circ}$]{
    \begin{minipage}[t]{0.19\linewidth}
    \centering
    \includegraphics[width=3.4cm]{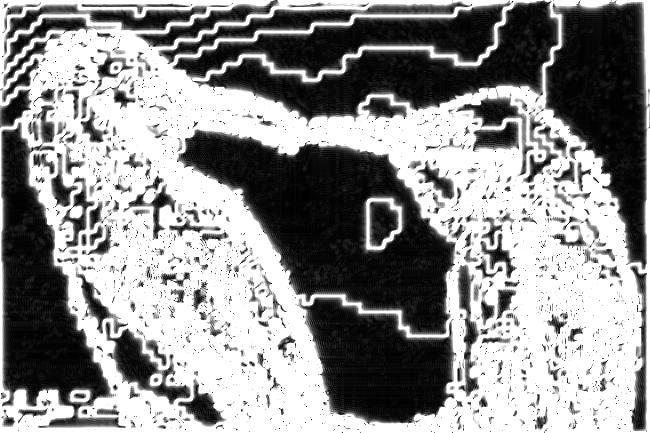}
    \end{minipage}%
    }%
    \subfigure[$90^{\circ}$]{
    \begin{minipage}[t]{0.19\linewidth}
    \centering
    \includegraphics[width=3.4cm]{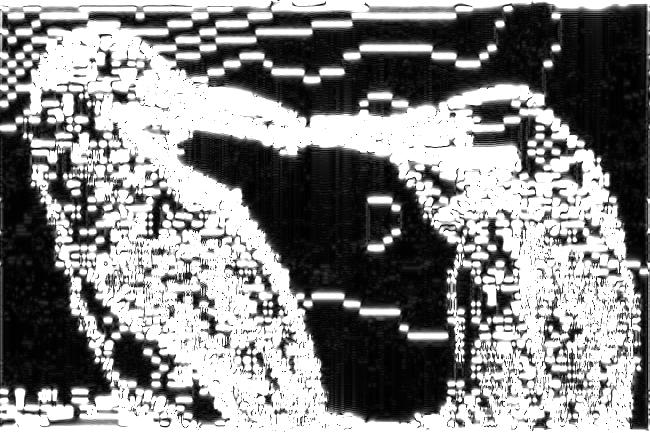}
    \end{minipage}%
    }%
    \subfigure[$135^{\circ}$]{
    \begin{minipage}[t]{0.19\linewidth}
    \centering
    \includegraphics[width=3.4cm]{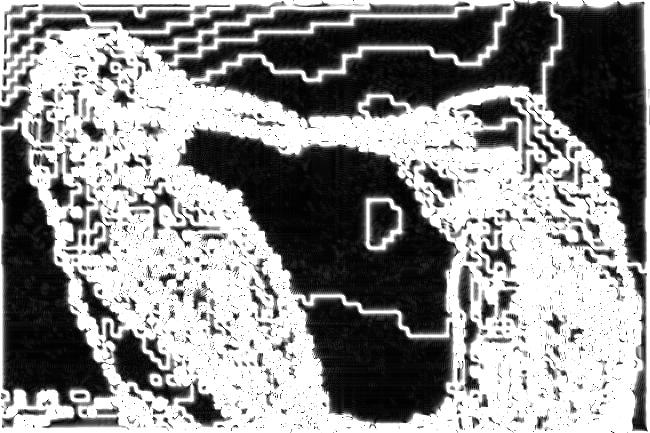}
    \end{minipage}
    }%
    
    \caption{An illustration of the reference image and two distorted images of bit rate b1 along with their Log-Gabor maps of 4 different orientations at frequency scale $\frac{2}{3}f_{0}$. Specifically, (f) is the compressed image with the uniform matrix as the optimization principle while (k) is the compressed image with MS-SSIM \cite{ms-ssim} as the optimization principle respectively.}
    \label{fig:texture}
\end{figure*}

\section{Proposed Method}
\label{sec:proposed}
\vspace{0.2cm}
The framework of the proposed method is clearly shown in Fig. \ref{fig:framework}. It can be seen that the proposed method first extracts features from gradient and texture information, and the final quality score is then computed by incorporating the similarities of gradient and texture feature maps. The gradient-based features and texture-based features exactly correspond to the process of perceiving compressed image quality from coarse and fine levels. Specifically, the gradient features can help reflect obvious distortions \cite{gsmd,dehaze1} such as structural damage and block effect while texture information is sensitive to the tiny compression artifacts \cite{texture1,texture2} such as texture shift.

\subsection{Pre-processing}
Since in most compression algorithms such as JPEG and JPEG2000, the image is first converted from $RGB$ color space to $YCbCr$ color space, where $Y$ indicates the luminance component, and $Cb$ and $Cr$ indicate the blue-difference and red-difference chroma components respectively. Thus, the same color space transformation is utilized as the pre-processing, which can be denoted as:
\begin{equation}
\left\{  
\begin{aligned}
Y&=0.257\cdot {R}+0.564\cdot {G}+0.098\cdot{B}+16, \\
Cb&=-0.148\cdot {R}-0.291\cdot {G}+0.439\cdot {B}+128, \\
{Cr}&=0.439\cdot {R}-0.368\cdot {G}-0.071\cdot{B}+128,
\end{aligned}\right.
\end{equation}
where $R,G,B$ represent the corresponding $RGB$ color channels, and we refer to the $Y,Cb,Cr$ color channels of the reference and distorted images as $Y_{r},Cb_{r},Cr_{r},Y_{d},Cb_{d},Cr_{d}$.

\subsection{Gradient-based Features}
\label{structure-based}
The gradient information has been proven to be very effective in many IQA methods \cite{gsmd} \cite{dehaze1}. As can be seen in Fig. \ref{fig:gradient}, the gradient maps can effectively reflect the artifacts caused by compression such as block damage and structural damage. These distortions are obvious to identify and can be easily quantified to function as the benchmark perception of fine-grained compressed images. To emphasize the importance of regions that are more sensitive to compression distortion, we employ the statistical parameters of the gradient similarity map in such regions for evaluation.

Given the reference luminance channel  $Y_{r}$ and the distorted one $Y_{d}$, a Sobel gradient operator is first applied to calculate the gradient maps. Then the gradient magnitude maps can be computed as:
\begin{equation}
\begin{array}{l}
\boldsymbol{G}_{r}=\sqrt{\left(Y_{r} \otimes {S}_{x}\right)^{2}+\left(Y_{r} \otimes {S}_{y}\right)^{2}}, \\
\boldsymbol{G}_{d}=\sqrt{\left(Y_{d} \otimes {S}_{x}\right)^{2}+\left(Y_{d} \otimes {S}_{y}\right)^{2}},
\end{array}
\end{equation}
where $\boldsymbol{G}_{r}$ and $\boldsymbol{G}_{d}$ represents the gradient magnitude maps of $Y_{r}$ and $Y_{d}$, the symbol $\otimes$ denotes the convolution operation, and ${S}_{x}$ and ${S}_{y}$ are the horizontal and vertical Sobel operators, which are defined as:
\begin{equation}
\boldsymbol{S}_{x}=\left[\begin{array}{ccc}
-\frac{1}{4} & 0 & -\frac{1}{4} \\
-\frac{1}{2} & 0 & -\frac{1}{2}\\
-\frac{1}{4} & 0 & -\frac{1}{4}
\end{array}\right], \boldsymbol{S}_{y}=\left[\begin{array}{ccc}
-\frac{1}{4} & -\frac{1}{2} & -\frac{1}{4} \\
0 & 0 & 0 \\
-\frac{1}{4} & -\frac{1}{2} & -\frac{1}{4}
\end{array}\right].
\end{equation}
With the computed gradient magnitude maps, the gradient similarity map $S_{g}$ can be simply calculated as:

\begin{equation}
    \boldsymbol{S_{g}} = \frac{2 {G}_{r} \cdot {G}_{d} + c_{1}}{{G}_{r}^2 + {G}_{d}^2 + c_{1}},
\end{equation}
where $c_{1}$ is a small constant value to avoid instability.
Fig. \ref{fig:texture} illustrates two typical kinds of areas where structure distortion usually occurs: one is the high-frequency area where structural damage and information loss may be easily perceived; the other is the relatively low-frequency area where the unexpected block and the structure over-restoration may be found. Specifically, the area can be represented as:

\begin{equation}
\begin{aligned}
    &\phi_{1}\! =  \! \{(i, j) \!  \mid  \! {G}_{r}(i, j )\!>\!E\left({G}_{r}\right) \bigcup  {G}_{d}(i, j)\! > \!E\left({G}_{d}\right) \},\\
    &\phi_{2} \! =  \!\{(i, j)\! \mid \! {G}_{d}(i, j)-{G}_{r}(i, j)\! >  \!E\left({G}_{d}-{G}_{r}\right)  \\ 
    & \qquad \qquad \qquad \qquad \bigcap  {G}_{r}(i,j) \!<\! E\left({G}_{r}\right) \},
\end{aligned}
\end{equation}
where $\phi_{1}$ represents the high-frequency area in which the pixel's gradient of the reference image or distorted image is larger than the average value of the corresponding gradient maps ${G}_{r}$ or ${G}_{d}$, and $\phi_{2}$ represents the low-frequency area in which the gradient difference of a pixel is larger than the average gradient difference and the pixel's gradient of the reference image is smaller than the average value of gradient map $G_{r}$.
Thus the mean and standard deviation of the gradient similarity maps in such areas are employed as the final features, which can be derived as:

\begin{equation}
\begin{aligned}
&\boldsymbol{E}_{g}=\frac{1}{T_{\phi}} \sum_{(i, j) \in \Phi} \boldsymbol{S}_{g}(i, j),\\
&\boldsymbol{Std}_{g}=\sqrt{\frac{1}{T_{\phi}} \sum_{(i, j) \in \Phi} (\boldsymbol{S}_{g}(i, j)-\boldsymbol{E}_{g})^2},\\
&\boldsymbol{\phi} = \phi_{1} \bigcup \phi_{2},
\end{aligned}
\end{equation}
where $\boldsymbol{E}_{g}$ and $\boldsymbol{Std}_{g}$ denote the mean and the standard deviation of the gradient similarity maps in area $\phi$, area $\phi$ is the union of $\phi_{1}$ and $\phi_{2}$, $T_{\phi}$ indicates the number of the pixels in $\phi$, and $(i,j)$ are the indices of the image pixels.



\subsection{Texture-based Features}
With the coarse perception features discussed above, we focus on the texture-based features in this section. For the quality assessment of compressed images, the texture is also a significant aspect, especially when the quality differences are subtle. Hence, we propose to use the Log-Gabor filter as the texture-based features extractor. The Log-Gabor filter is specially introduced to deal with the feature analysis \cite{Log-Gabor} and more importantly, it can determine whether there is any certain frequency content in specific directions in the localized region of an image. Fig. \ref{fig:texture} shows the illustration of fine-grained compressed images with uniform matrix and MS-SSIM as the optimization principle. Although the two compressed images share almost the same bit rate, the texture maps vary from each other. Different texture shifts can be observed among all the four orientations' texture maps, which indicates that texture information has a good ability to catch the slight artifacts of compression. Then, the reference and compressed images are first decomposed through a 2-D oriented Log-Gabor transformation, which can be derived as:

\begin{equation}
\boldsymbol{G(f, \theta)}\!=\!\exp \left(\frac{-\left(\log \left(f / f_{0}\right)\right)^{2}}{2\left(\log \left(\sigma_{f} / f_{0}\right)\right)^{2}}\right) \!\cdot \exp\! \left(\!\frac{-\left(\theta-\theta_{0}\right)^{2}}{2 \sigma_{\theta}^{2}}\!\right)\!,
\vspace{0.3cm}
\end{equation}
where $f$ and $\theta$ indicate the frequency and orientation of texture, $f_{0}$ and $\theta_{0}$ denote the center frequency and center orientation, and $\delta_{f}$ and $\delta_{\theta}$ are width parameters for frequency and orientation. In this paper, parameters mentioned above are set to cover five frequency scales ($\frac{2}{3}f_{0}$,  $\frac{4}{3}f_{0}$, $2f_{0}$, $\frac{8}{3}f_{0}$, $\frac{10}{3}f_{0}$) with each band spanning an approximately $\frac{2}{3}f_{0}$ bandwidth. These frequency values are chosen to empirically reduce the effect of overlap between bands and to cover the major frequency spectrum that counts for human vision system \cite{mad}.  Four orientations ($0^{\circ}, 45^{\circ},90^{\circ},135^{\circ}$) of texture are included in experiment, which tends to be a valid setup for natural images.

All of the $YCbCr$ color channels of the reference and compressed images are filtered through the Log-Gabor filter with parameters mentioned above, which means that there are a total of 120 $=5\times4\times2\times3$ (frequency scales $\times$ orientations $\times$ reference and compressed images $\times$ $YCbCr$ color channels) Log-Gabor maps computed. Since the Log-Gabor transformation results consist of real and imaginary parts, we utilize the amplitude of the Log-Gabor maps of reference and distorted images with all $YCbCr$ channels as the feature maps, which can be denoted as $Y_{ra}$, $Cb_{ra}$, $Cr_{ra}$, $Y_{da}$, $Cb_{da}$, and $Cr_{da}$ respectively. Therefore, the weighted average sums of similarity maps for $YCbCr$ color space's Log-Gabor results are computed as the texture similarity map:

\begin{equation}
\begin{aligned}
    \boldsymbol{T_{\Lambda}} =  \sum_{f=1}^{5}   W_{ga}  & \cdot  \sum_{o=1}^{4}  \frac{2 {\Lambda}_{ra}(f,o) \cdot {\Lambda}_{da}(f,o) + c_{2}}{{\Lambda}_{ra}^2(f,o) + {\Lambda}_{da}^2(f,o) + c_{2}}, \\
    \Lambda & \in \{Y,Cb,Cr\},\\
\end{aligned}
\label{equ:wa}
\end{equation}
where $\boldsymbol{T}_{\Lambda}$ denotes the texture similarity map for $YCbCr$ channels, $\Lambda$ is the indicator of specific $YCbCr$ channels, $c_{2}$ is a small constant value to avoid instability. $W_{ga}$ stands for the weight for each frequency component.  Then the final texture similarity map can be computed as:

\begin{equation}
    \boldsymbol{S_{t}} = \sqrt{W_{Y} \cdot T_{Y}^2 + W_{Cb} \cdot  \frac{1}{4} T_{Cb}^2 +  W_{Cr} \cdot  \frac{1}{4} T_{Cr}^2},
    \label{equ:st}
\end{equation}
where $W_{Y}$, $W_{Cb}$, and $W_{Cr}$ indicate the weight for $T_{Y}$, $T_{Cb}$, and $T_{Cr}$ components respectively. Finally, the texture features can be computed as the mean and standard deviation of the texture similarity map:

\begin{equation}
\begin{aligned}
&\boldsymbol{E}_{t}=\frac{1}{T} \sum \boldsymbol{S}_{t}(i, j),\\ \vspace{-0.1cm}
&\boldsymbol{Std}_{t}=\sqrt{\frac{1}{T} \sum (\boldsymbol{S}_{t}(i, j)-\boldsymbol{E}_{t})^2},\\
\end{aligned}
\label{equ:parameter}
\end{equation}
where $\boldsymbol{S}_{t}$ stands for the texture similarity map, $\boldsymbol{E}_{t}$ and $\boldsymbol{Std}_{t}$ are the mean and standard deviation of the texture similarity map, $T$ indicates the number of the pixels in the image, and $(i,j)$ are the indices of the pixels.

\begin{table*}[t]
\renewcommand\arraystretch{1.5}
\caption{Validated results on the FGIQA database.}
\centering
\begin{tabular}{l|c|c|c|c|c|c|c|c|c|c|c|c}
\toprule
\multicolumn{1}{l}{\multirow{2}{*}{Mehtods}} & \multicolumn{4}{|c}{SRCC}                                              & \multicolumn{4}{|c}{KRCC}                                              & \multicolumn{4}{|c}{PLCC}                                              \\ \cline{2-13}
\multicolumn{1}{c|}{}                        & b1              & b2              & b3              & average         & b1              & b2              & b3              & average         & b1              & b2              & b3              & average         \\ \hline
PSNR                                        & 0.5021          & -0.6857         & -0.8373         & -0.3403         & 0.4467          & -0.5661         & -0.7702         & -0.2965         & 0.5279          & -0.6777         & -0.8498         & -0.3332         \\
SSIM                                        & 0.6362          & 0.2194          & -0.4006         & 0.1517          & 0.5412          & 0.2203          & -0.2647         & 0.1656          & 0.8086          & 0.8515          & 0.8004          & 0.8202          \\
VSI                                         & 0.7885          & 0.7825          & 0.6891          & 0.7534          & 0.6974          & 0.6991          & 0.6368          & 0.6778          & 0.8081          & 0.8928          & 0.6998          & 0.8002          \\
MS-SSIM                                     & 0.8965          & \textbf{0.8649}          & 0.7959          & \textbf{0.8524} & 0.8607          & 0.8098          & 0.7432          & \textbf{0.8046} & \textbf{0.9157} & \textbf{0.9163} & 0.8420          & \textbf{0.8913} \\
IW-SSIM                                      & 0.7467          & 0.8644          & \textbf{0.8771} & 0.7237          & 0.7933          & \textbf{0.8107} & 0.6361          & 0.8217          & 0.7851          & 0.6545          & 0.8183          & 0.7526          \\
FSIM                                        & 0.7545          & 0.7872          & 0.3999          & 0.6472          & 0.6907          & 0.7042          & 0.5498          & 0.6482          & 0.8185          & 0.8780          & 0.7135          & 0.8034          \\
MAD                                         & \textbf{0.9352} & 0.8295          & 0.6169          & 0.7938          & \textbf{0.8765} & 0.7248          & 0.5528          & 0.7180          & 0.9074          & 0.8286          & 0.6031          & 0.7797          \\
GMSD                                        & 0.7845          & 0.8029          & 0.8659 & 0.8178          & 0.6907          & 0.7164          & \textbf{0.8198} & 0.7423          & 0.8313          & 0.9097          & \textbf{0.8727} & 0.8712          \\
IFC                                         & 0.8113          & 0.8612          & 0.8047          & 0.8257          & 0.7344          & 0.6945          & 0.8011          & 0.7433          & 0.7894          & 0.8733          & 0.8577          & 0.8401          \\
PieAPP                                      & 0.7521          & 0.7941          & 0.7699          & 0.7721          & 0.6767          & 0.7101          & 0.6998          & 0.6956          & 0.7926          & 0.8666          & 0.8051          & 0.8241          \\
LPIPS                                       & 0.6875          & 0.7027          & 0.7100          & 0.7001          & 0.6391          & 0.6628          & 0.6634          & 0.6651          & 0.7068          & 0.7439          & 0.7300          & 0.7269          \\
Proposed                                    & \textbf{0.9505} & \textbf{0.8760} & \textbf{0.8671} & \textbf{0.8978} & \textbf{0.9174} & \textbf{0.8201} & \textbf{0.8035} & \textbf{0.8470} & \textbf{0.9551} & \textbf{0.9096} & \textbf{0.8749} & \textbf{0.9132}\\
\bottomrule
\end{tabular}
\label{tab:fgiqa}
\end{table*}

\begin{figure*}[h]
    \centering
    \subfigure[b1]{
    \begin{minipage}[t]{0.3\linewidth}
    \centering
    \includegraphics[width=5cm,height = 5.5cm]{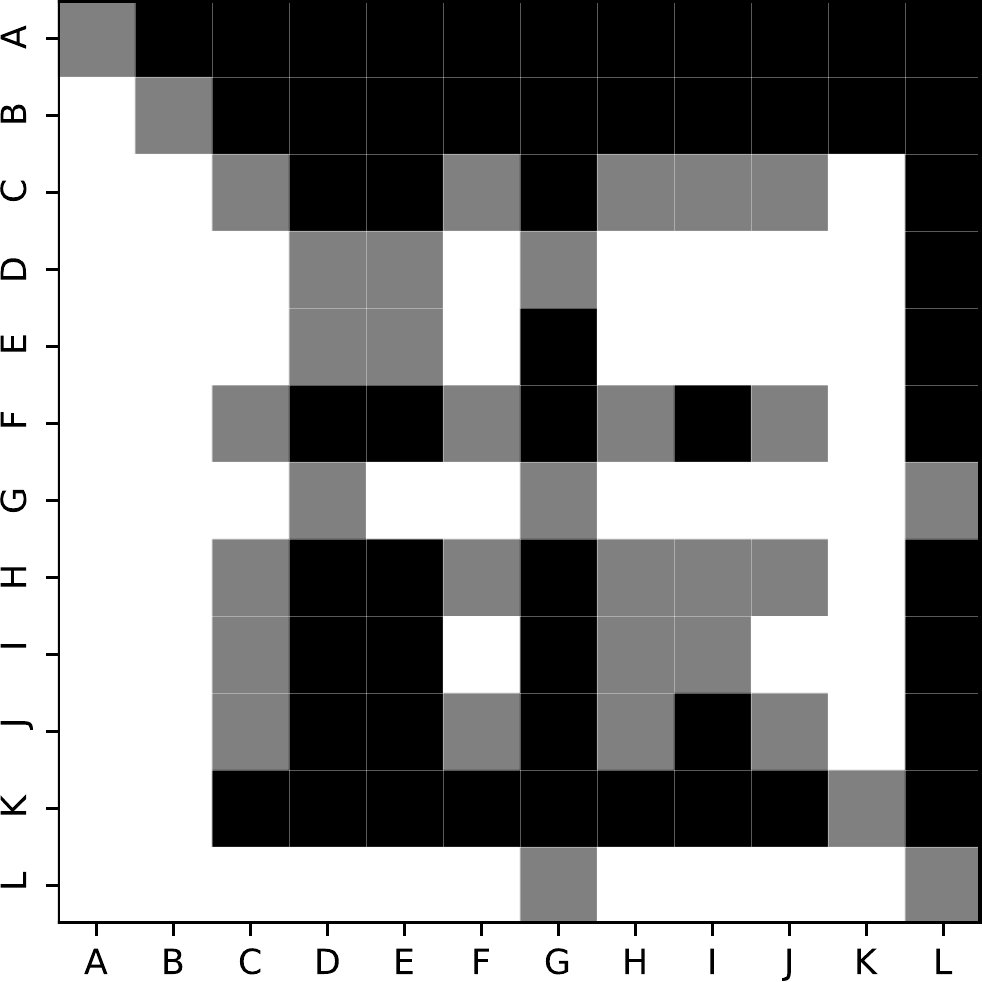}
    \end{minipage}%
    }%
    \subfigure[b2]{
    \begin{minipage}[t]{0.3\linewidth}
    \centering
    \includegraphics[width=5cm,height = 5.5cm]{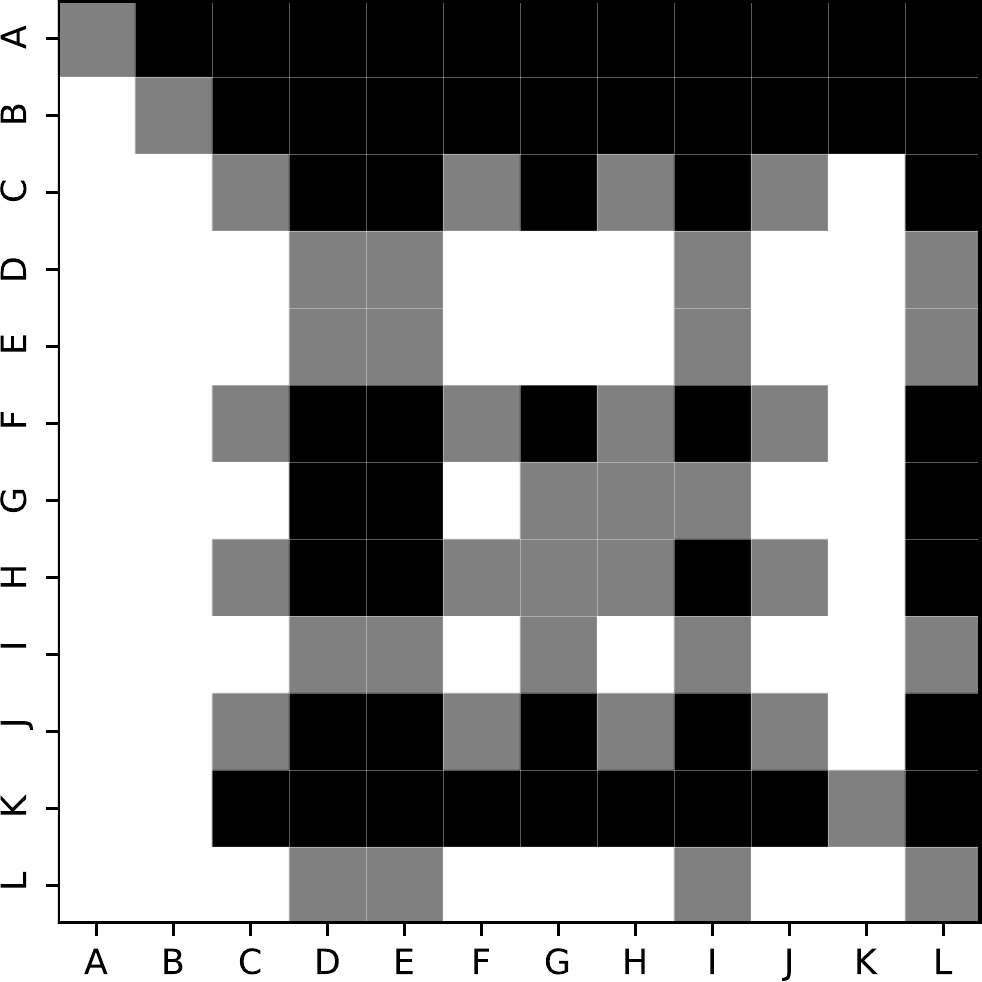}
    \end{minipage}%
    }%
    \subfigure[b3]{
    \begin{minipage}[t]{0.3\linewidth}
    \centering
    \includegraphics[width=5cm,height = 5.5cm]{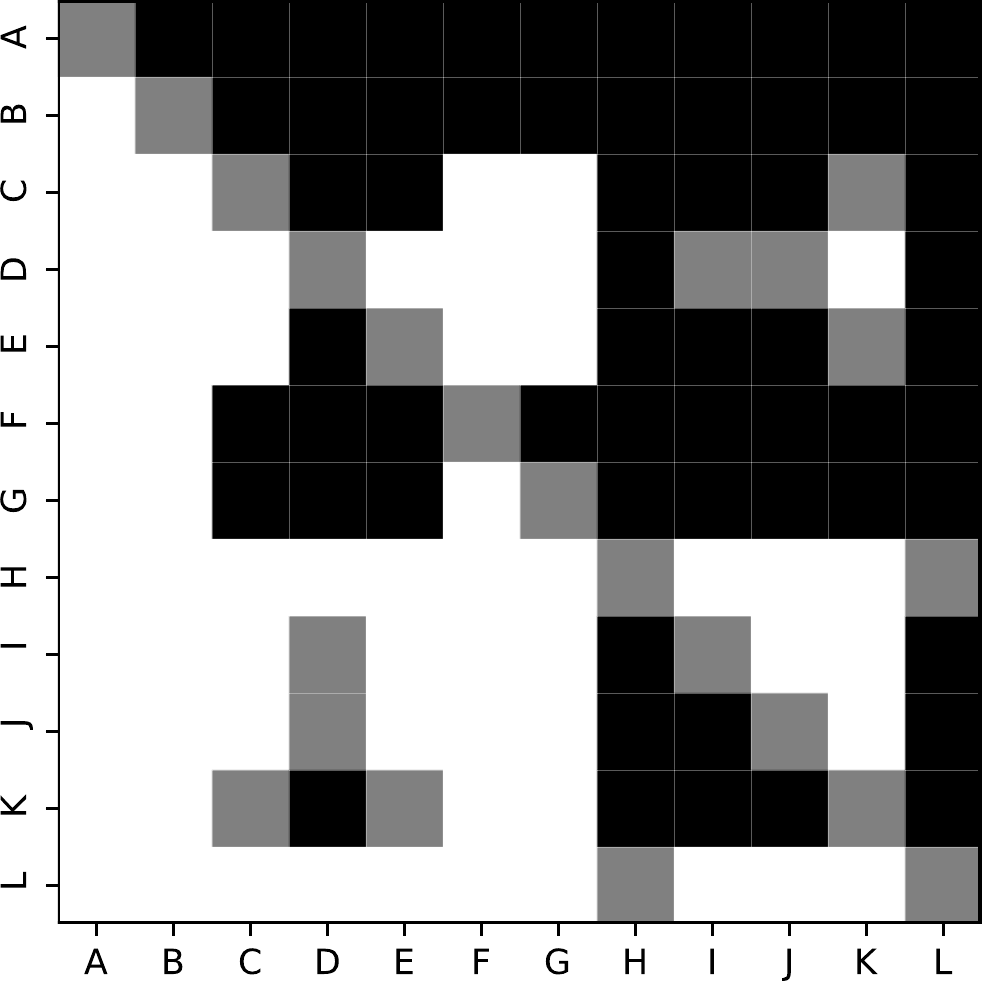}
    \end{minipage}%
    }%
    \caption{Statistical test results of the proposed method and compared methods on the FGIQA database. A black/white block means the row method is statistically worse/better than the column one. A gray block means the row method and the column method are statistically indistinguishable. The methods denoted by A-L are of the same order as in Table \ref{tab:fgiqa}. (a), (b), and (c) are the statistical test results for bit rates b1, b2, and b3 respectively.}
    \label{fig:heatmap}
\end{figure*}

\subsection{Overall Quality Score Computation}

After extracting the gradient-based and texture-based features, the final quality score can be computed using these features:

\begin{equation}
    \boldsymbol{Q} = \frac{E_{g}^{\alpha} \cdot E_{t}^{\beta}}{Std_{g}^{\alpha}\cdot Std_{t}^{\beta}} ,
    \vspace{0.3cm}
\end{equation}
where $\boldsymbol{Q}$ denotes the overall quality score, $E_{g}$, $E_{t}$, $Std_{g}$, and $Std_{t}$ stand for the extracted features described in the above sections, $\alpha$, and $\beta$ are the parameters to control the influence of each feature group. More specifically, $E_{g}$ and $E_{t}$ describe the mean value of corresponding similarity maps, which are positively correlated with the score. However, $Std_{g}$ and $Std_{t}$ describe the fluctuation of similarity, which are thus negatively correlated with the score.  To further test the influence of different values of $\alpha$ and $\beta$, we also conduct the parameter sensitivity experiment in Section \ref{sec:parameter}.

\begin{table*}[h]
\centering
\renewcommand\arraystretch{1.5}
\caption{Performance results on other compression databases.}
\begin{tabular}{c|l|cc|cc|cc|c}
\toprule
\multicolumn{1}{l|}{\multirow{2}{*}{}} &\multicolumn{1}{l}{\multirow{2}{*}{Methods}} &\multicolumn{2}{|c}{LIVE} &\multicolumn{2}{|c}{TID2013} &\multicolumn{2}{|c|}{CSIQ} &\multicolumn{1}{l}{\multirow{2}{*}{Average}}
\\ \cline{3-8}
\multicolumn{1}{l|}{}  &    & JPEG & JP2K & JPEG & JP2K & JPEG & JP2K & \multicolumn{1}{c}{} \\ \hline
\multirow{7}{*}{SRCC} & SSIM     & 0.9327    & 0.8744    & 0.9150   & 0.9505   & 0.9583    & 0.9616    & 0.9321                      \\
                      & MS-SSIM  & 0.8746    & 0.8417    & 0.9096   & 0.9049   & 0.9219    & 0.9205    & 0.8955                      \\
                      & GMSD     & 0.9340    & 0.9097    & 0.9507   & 0.9657   & \textbf{0.9659}    & \textbf{0.9717}    & 0.9496                      \\
                      & VSI      & 0.9236    & 0.8856    & \textbf{0.9541}   & \textbf{0.9706}   & 0.9623    & 0.9692    & 0.9442                      \\
                      & MAD      & 0.9710    & 0.9411    & 0.9217   & 0.9511   & 0.9613    & 0.9752    & 0.9536                      \\
                      & PieAPP   & 0.9395    & 0.9056    & 0.8110   & 0.9417   & 0.9456    & 0.9447    & 0.9147                      \\
                      & Proposed & \textbf{0.9730}    & \textbf{0.9651}    & 0.9143   & 0.9439   & 0.9622    & 0.9685    & \textbf{0.9545}             \\\hline
\multirow{7}{*}{KRCC} & SSIM     & 0.7648    & 0.6864    & 0.7308   & 0.8062   & 0.8224    & 0.8268    & 0.7729                      \\
                      & MS-SSIM  & 0.6850    & 0.6562    & 0.7334   & 0.7202   & 0.7540    & 0.7532    & 0.7170                      \\
                      & GMSD     & 0.7646    & 0.7321    & 0.8036   & 0.8400   & \textbf{0.8387}    & 0.8525    & 0.8052                      \\
                      & VSI      & 0.7448    & 0.6963    & \textbf{0.8121}   & \textbf{0.8539}   & 0.8280    & 0.8473    & 0.7971                      \\
                      & MAD      & 0.8549    & 0.7852    & 0.7238   & 0.8139   & 0.8337    & \textbf{0.8699}    & 0.8136                      \\
                      & PieAPP   & 0.7824    & 0.7351    & 0.6417   & 0.7909   & 0.7843    & 0.7870    & 0.7535                      \\
                      & Proposed & \textbf{0.8573}    & \textbf{0.8389}    & 0.7192   & 0.7909   & 0.8280    & 0.8516    & \textbf{0.8143}             \\ \hline
\multirow{7}{*}{PLCC} & SSIM     & 0.8631    & 0.8142    & 0.9298   & 0.9506   & 0.9360    & 0.9106    & 0.8340                      \\
                      & MS-SSIM  & 0.9041    & 0.8753    & 0.9187   & 0.8950   & 0.9033    & 0.8777    & 0.8290                      \\
                      & GMSD     & 0.8595    & 0.9137    &\textbf{ 0.9736}   & \textbf{0.9788}   &0.9438    & 0.9633   & 0.8721                      \\
                      & VSI      & 0.9269    & 0.8908    & 0.9479   & 0.9494   & 0.8830    & 0.9008    & 0.8098                      \\
                      & MAD      & 0.8599    & 0.8929    & 0.9510   & 0.9663   & 0.9694    & 0.9807    & 0.9034                      \\
                      & PieAPP   & 0.8159    & 0.8246    & 0.7783   & 0.9027   & 0.9680    & 0.9545    & 0.8740                      \\
                      & Proposed & \textbf{0.9612}    & \textbf{0.9279}    & 0.9461   & 0.9305   & \textbf{0.9717}    & \textbf{0.9816}    & \textbf{0.9531}   \\
                      \bottomrule
\end{tabular}
\label{tab:others}
\end{table*}

\begin{figure*}[!htp]
    \centering
    \subfigure[]{\includegraphics[width = 5.13cm]{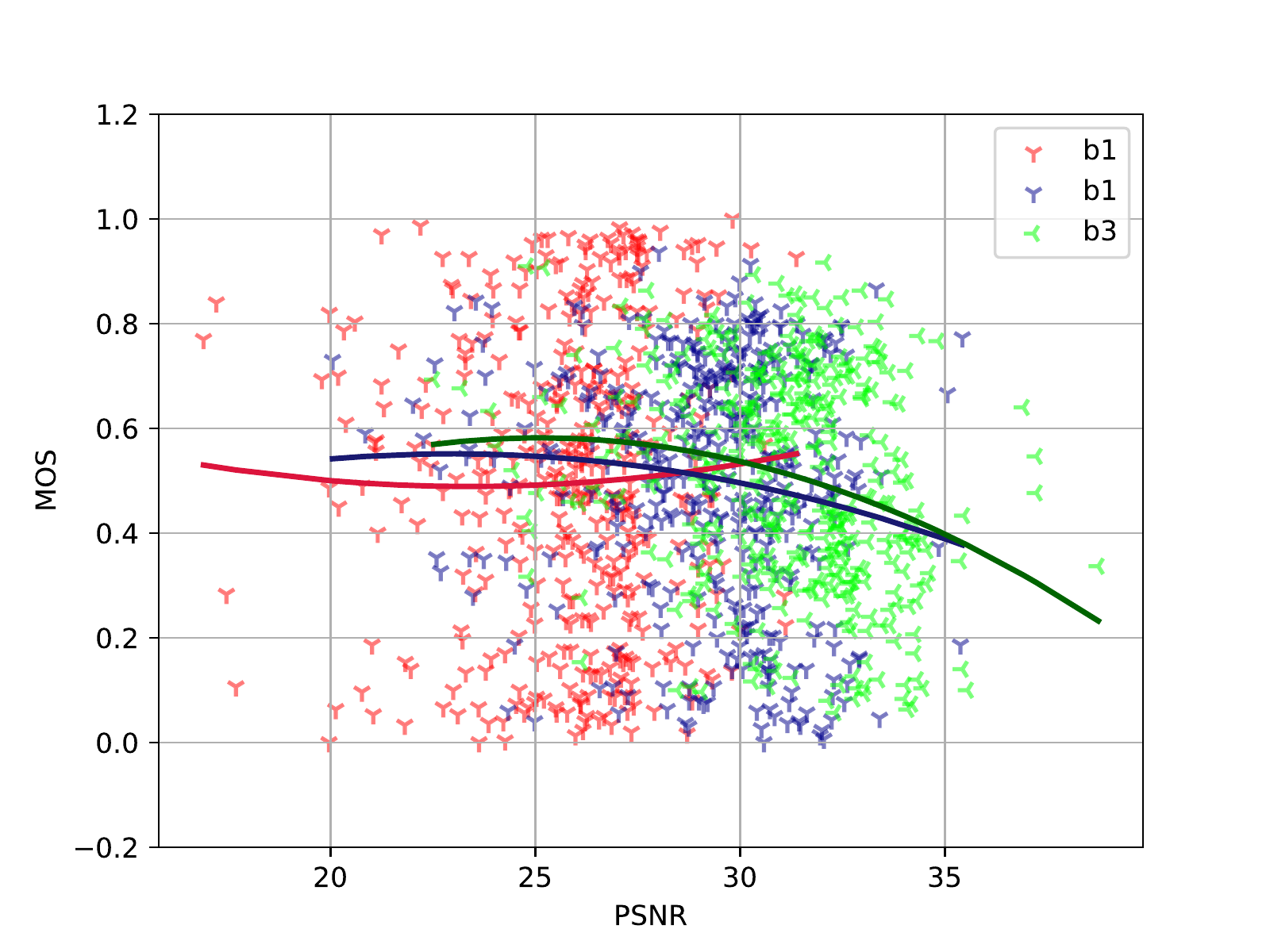}}
    \subfigure[]{\includegraphics[width = 5.13cm]{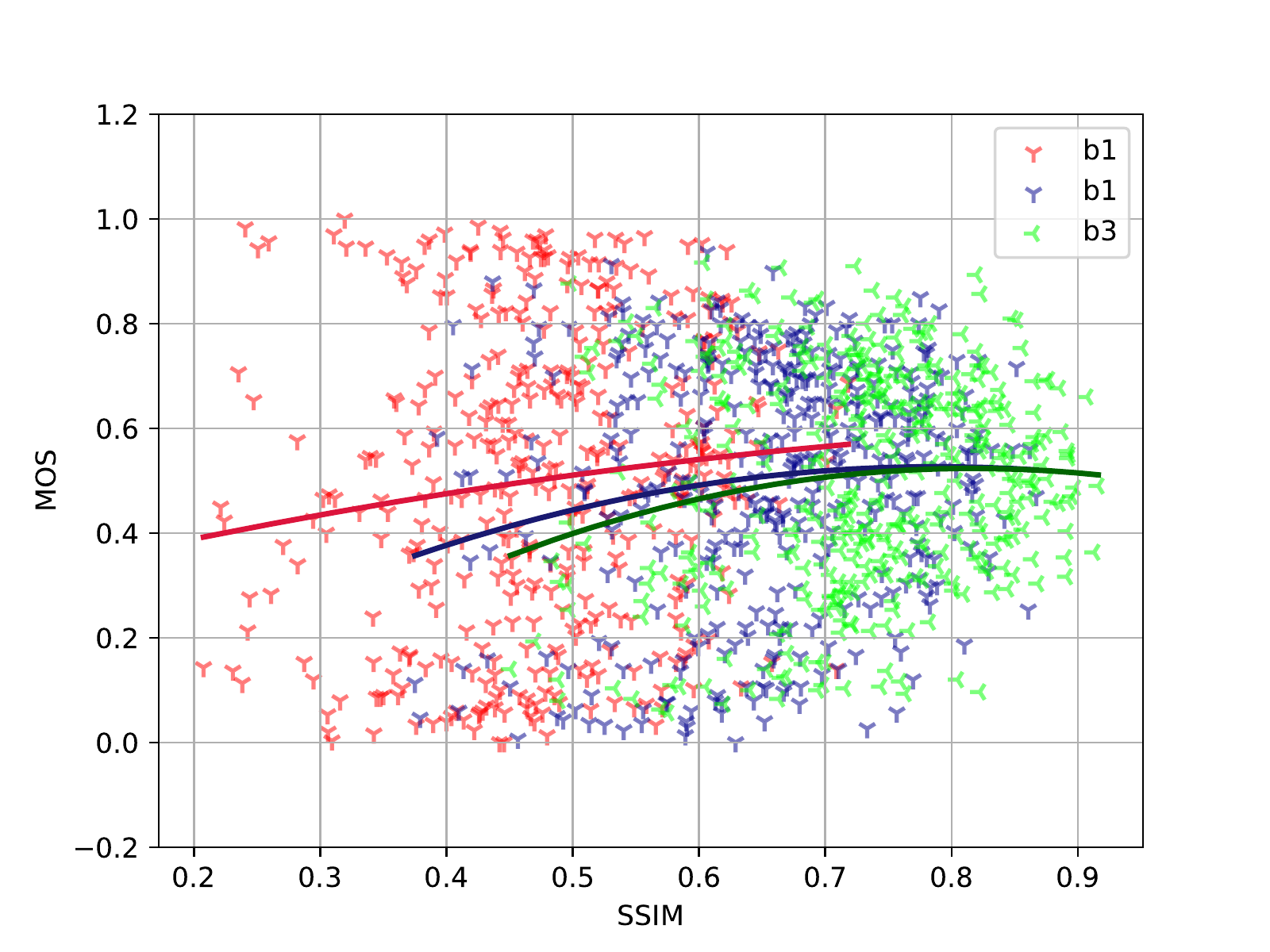}}
    \subfigure[]{\includegraphics[width = 5.13cm]{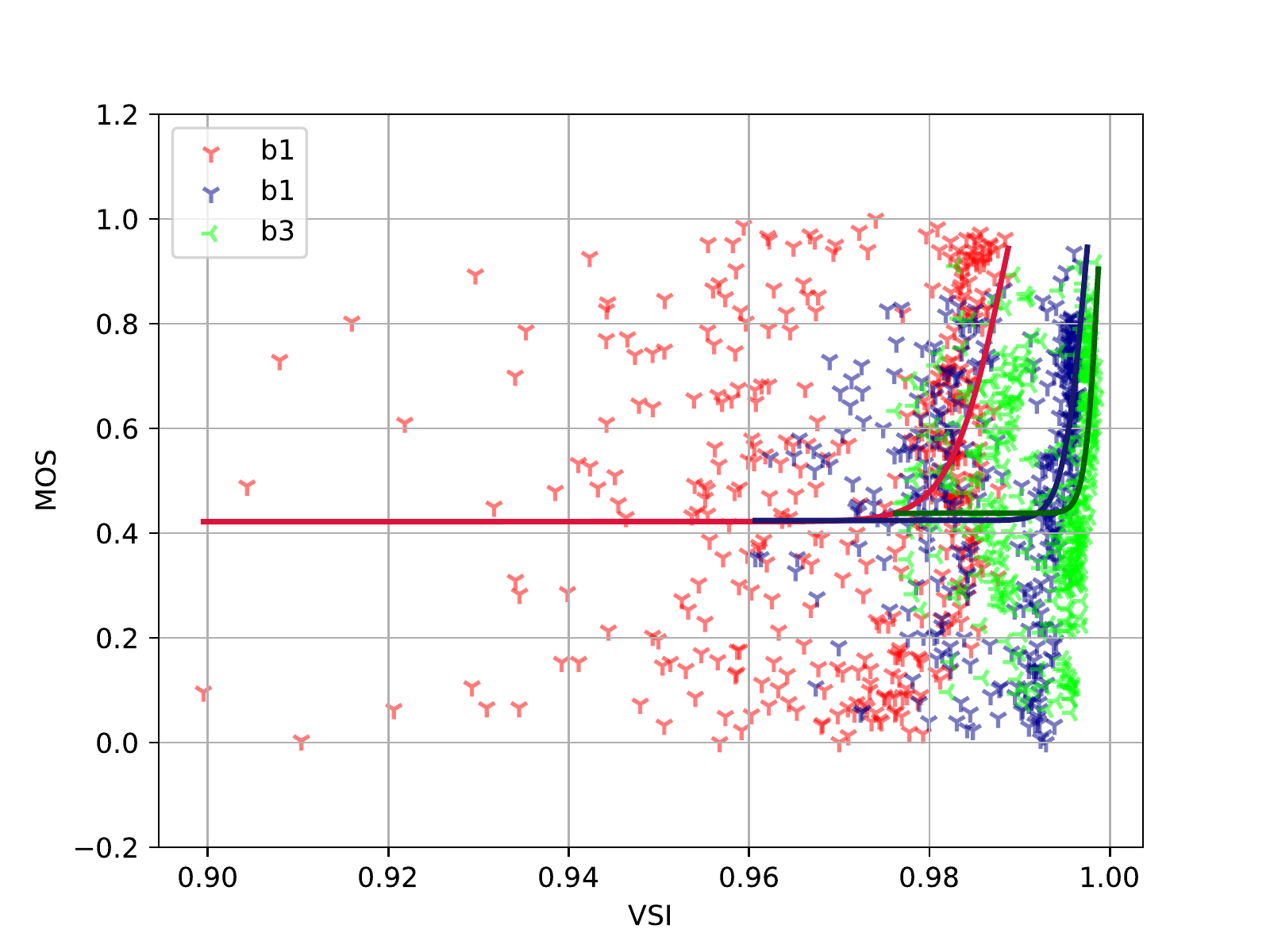}}
    \subfigure[]{\includegraphics[width = 5.13cm]{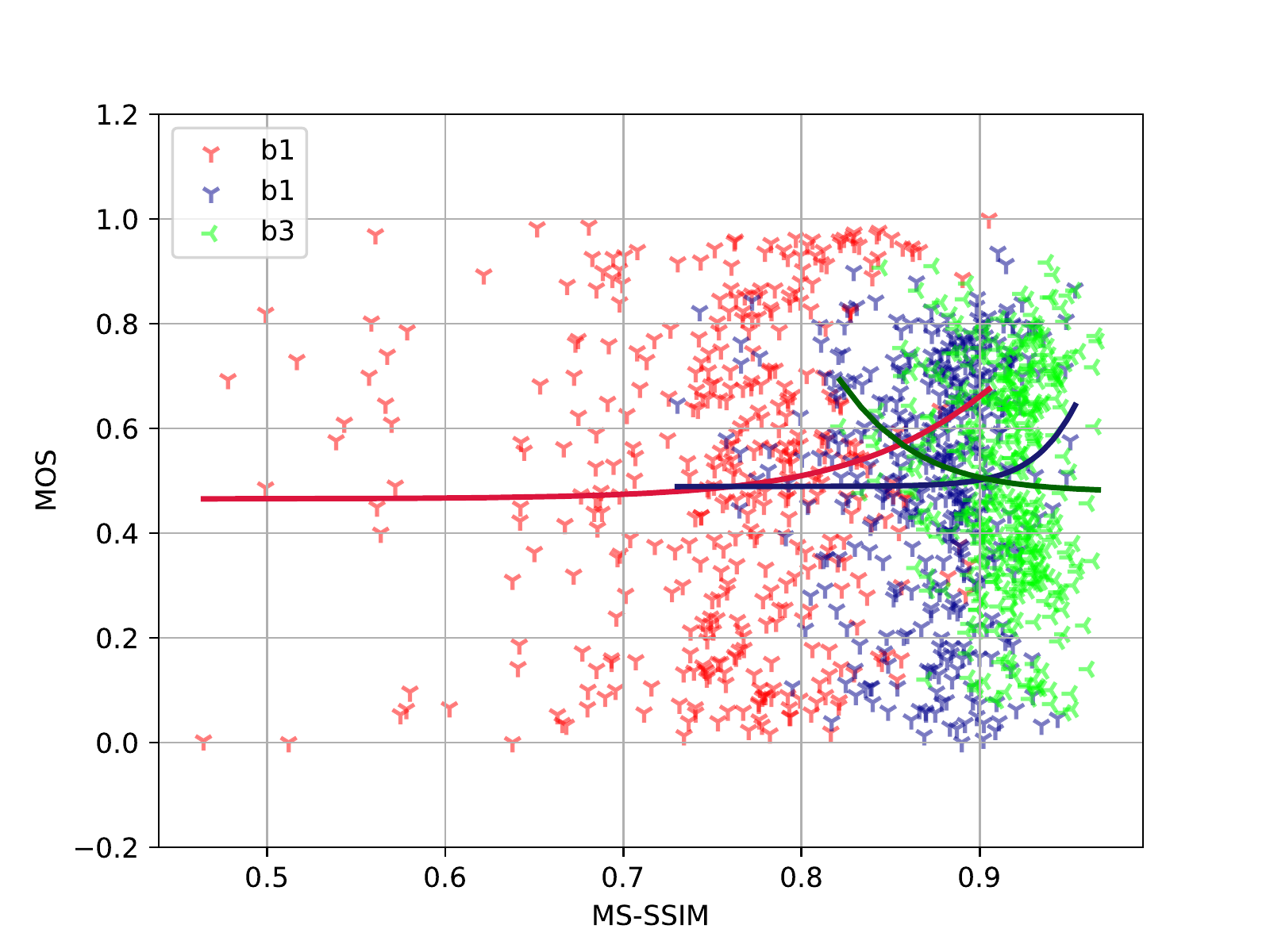}}
    \subfigure[]{\includegraphics[width = 5.13cm]{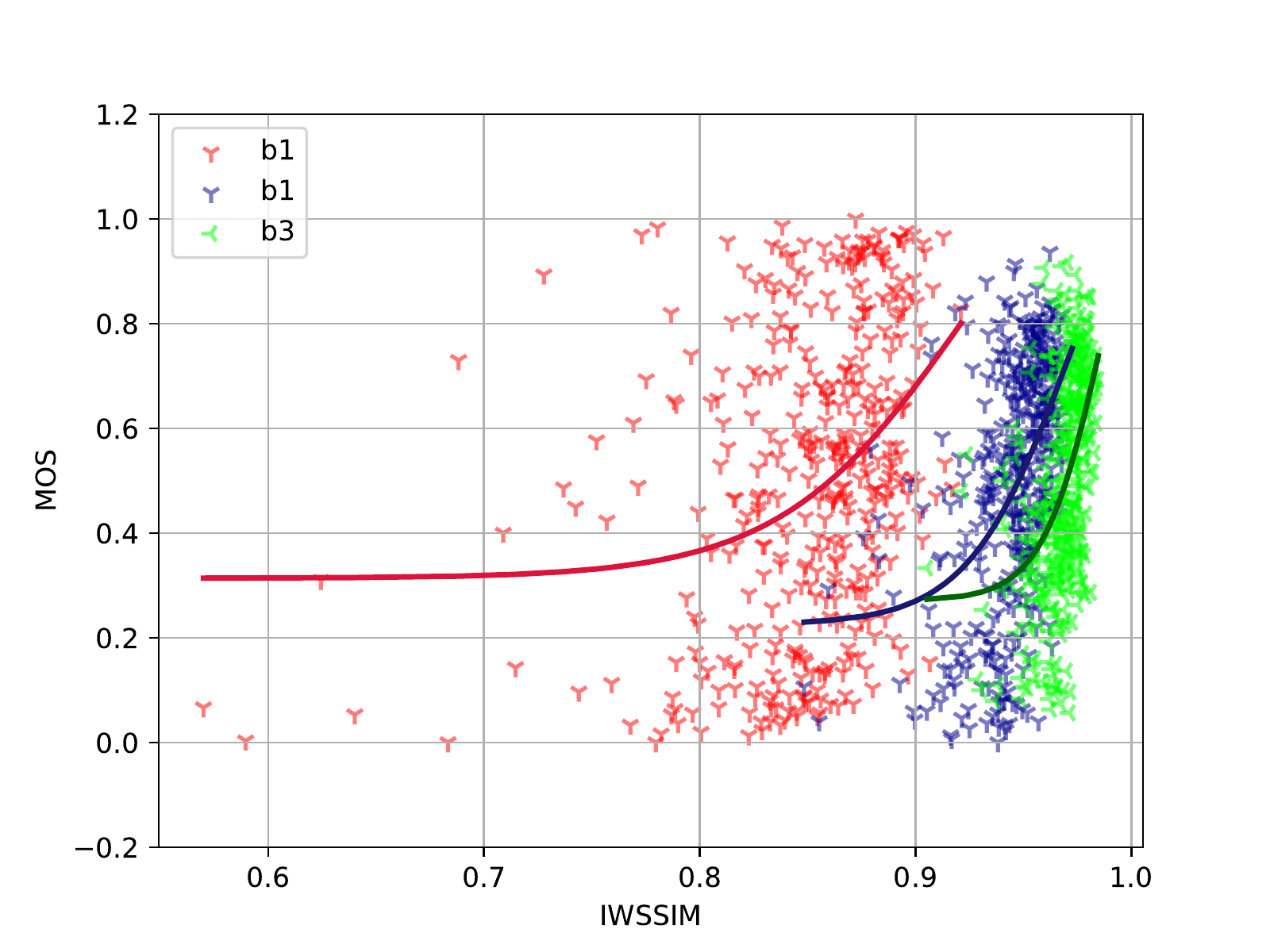}}
    \subfigure[]{\includegraphics[width = 5.13cm]{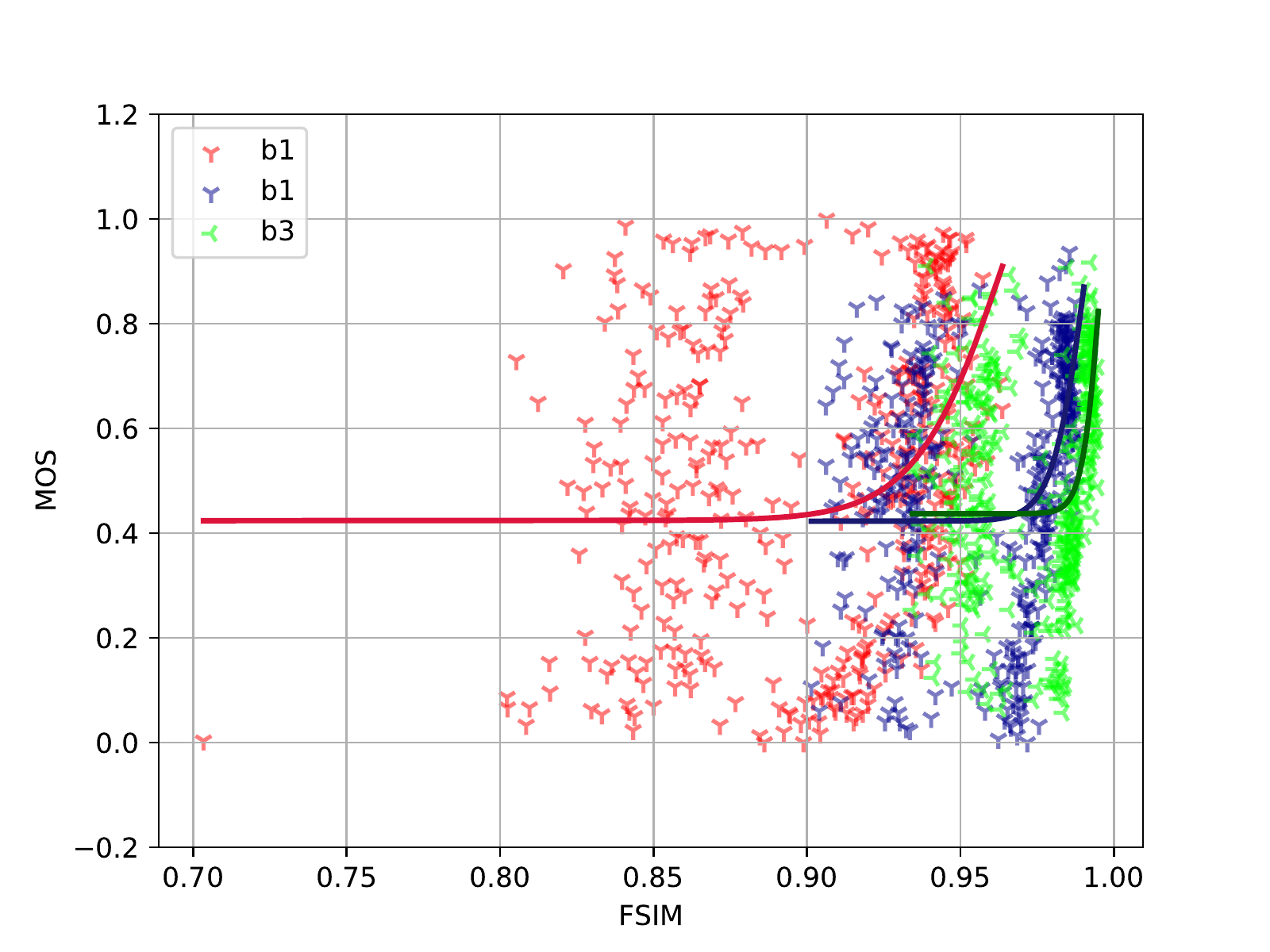}}
    \subfigure[]{\includegraphics[width = 5.13cm]{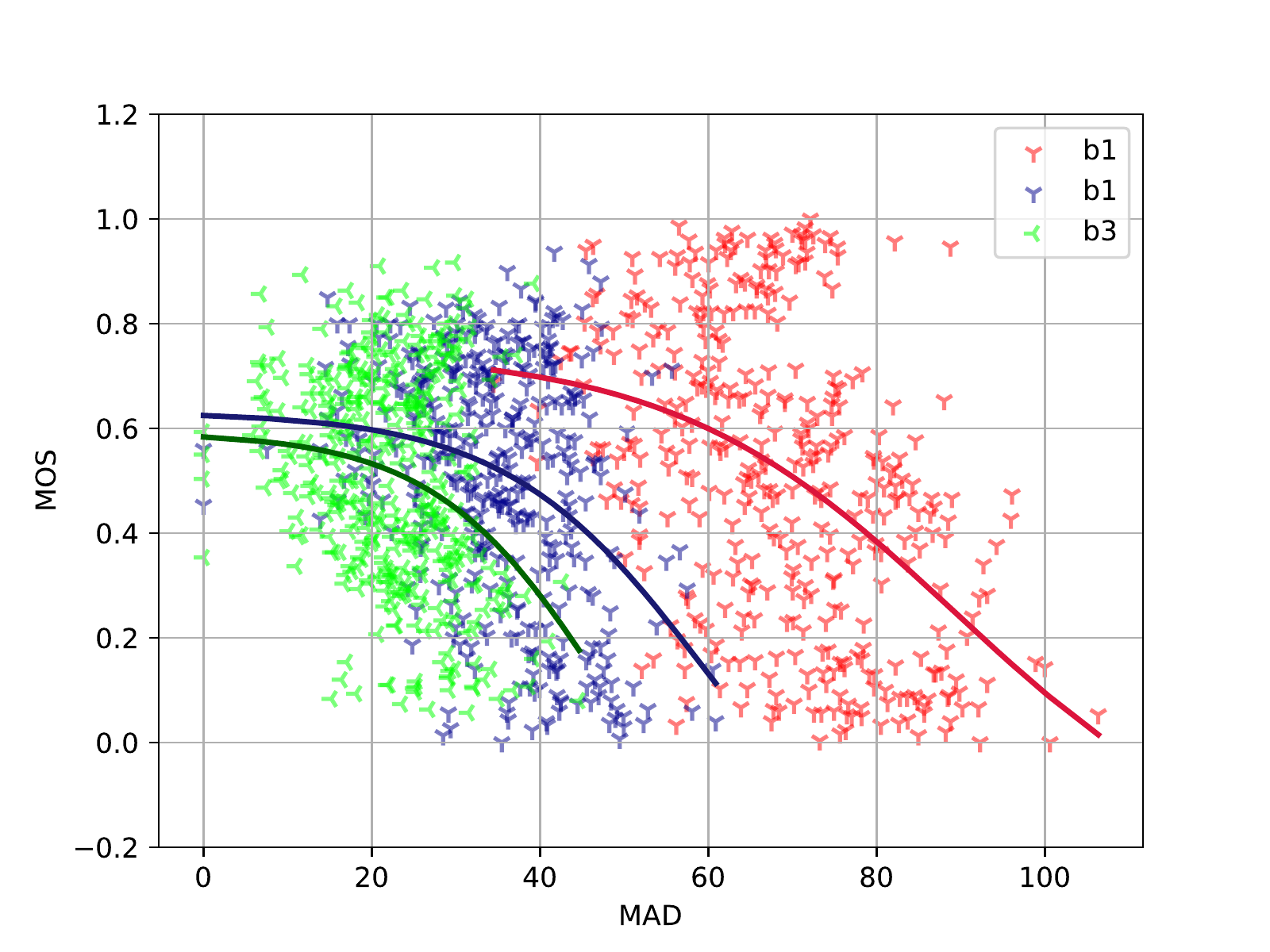}}
    \subfigure[]{\includegraphics[width = 5.13cm]{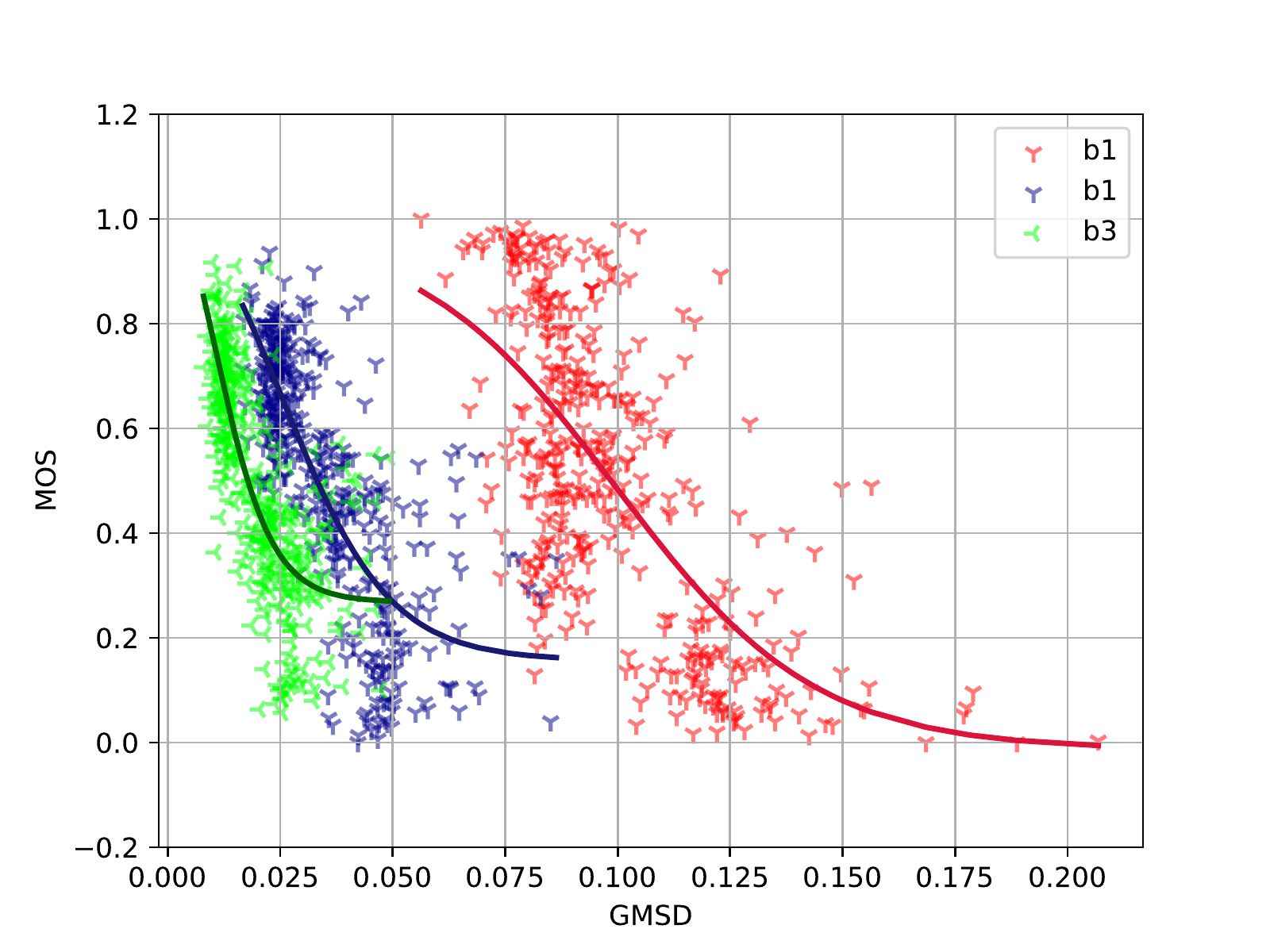}}
    \subfigure[]{\includegraphics[width = 5.13cm]{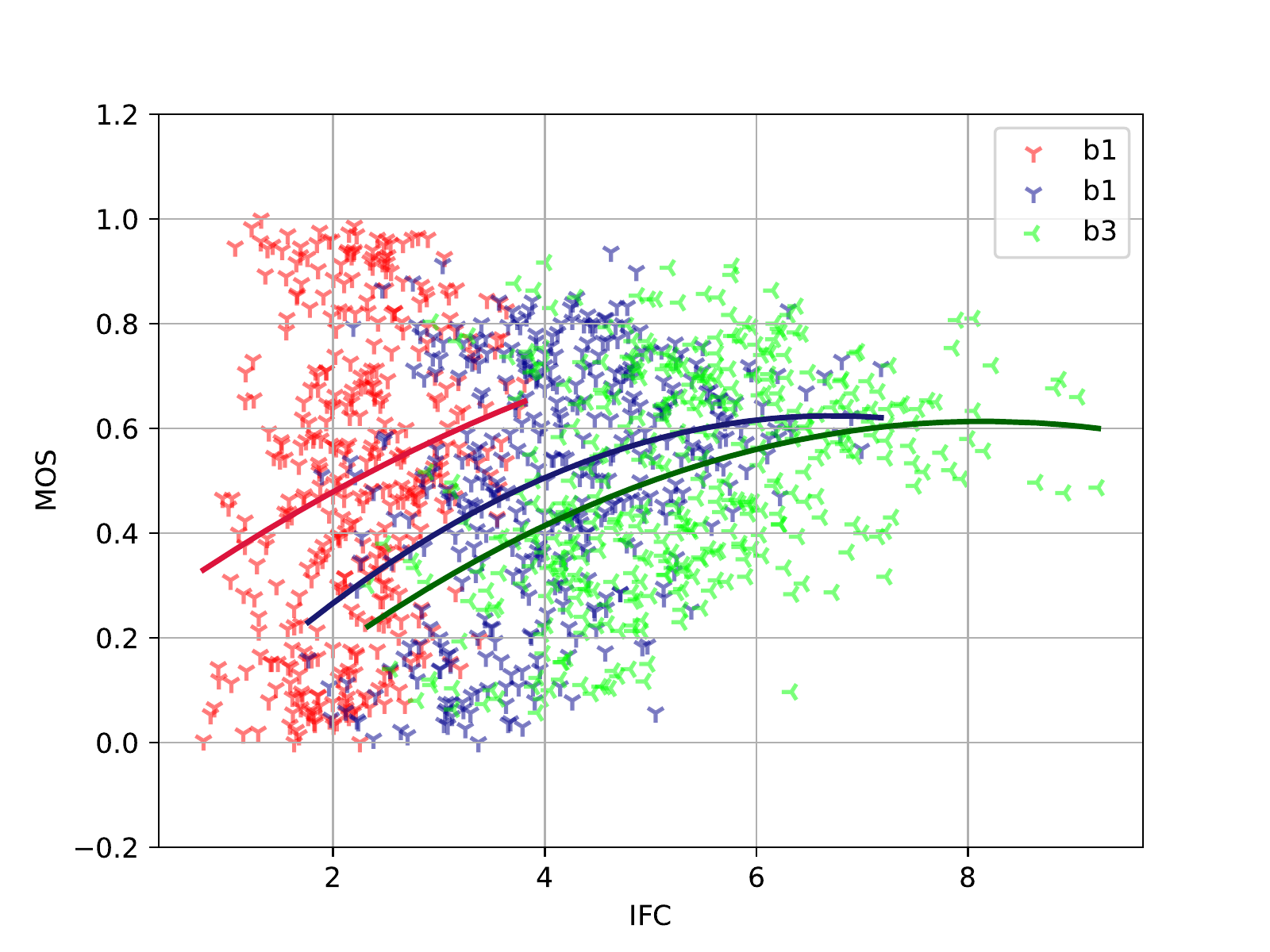}}
    \subfigure[]{\includegraphics[width = 5.13cm]{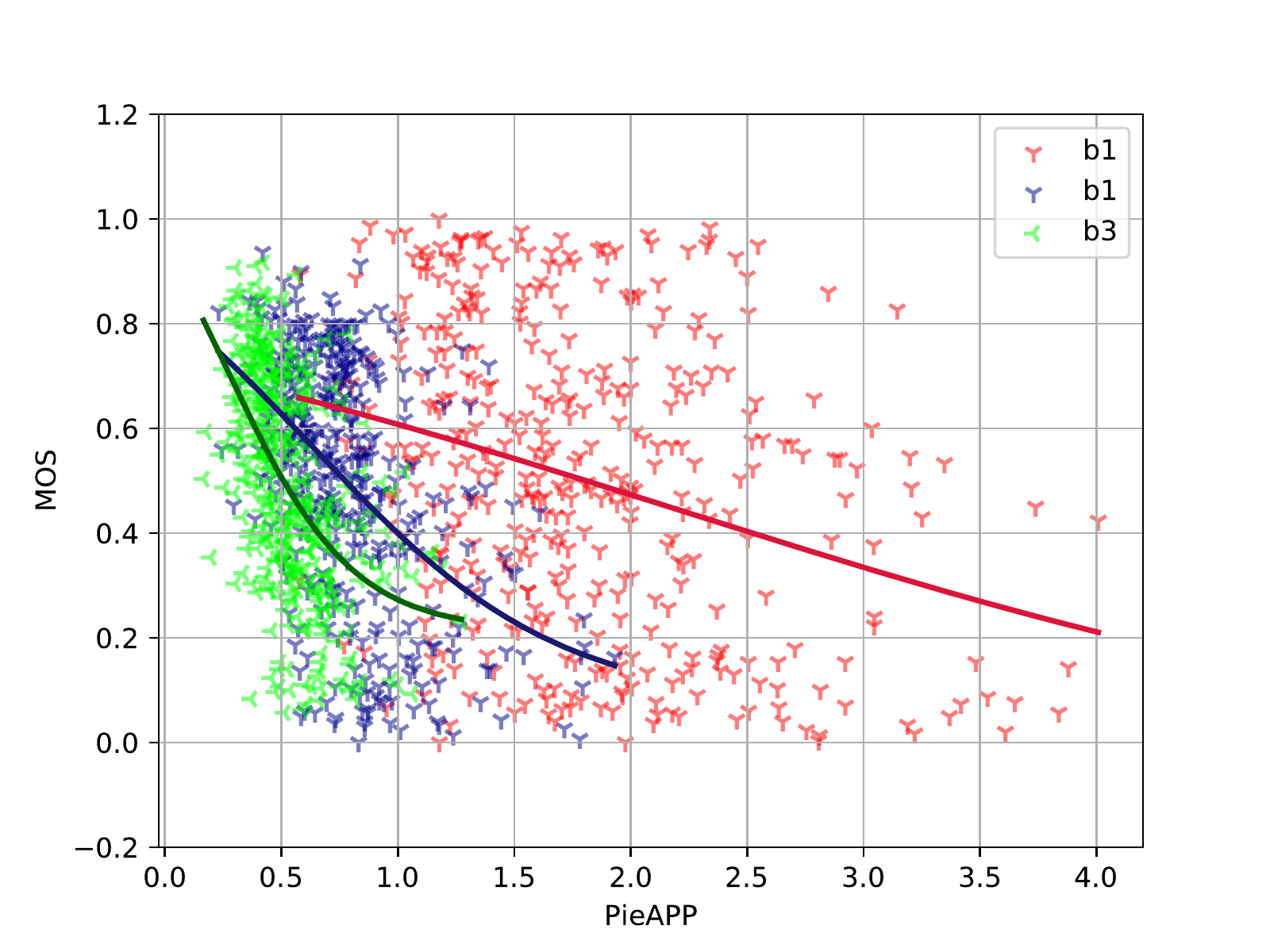}}
    \subfigure[]{\includegraphics[width = 5.13cm]{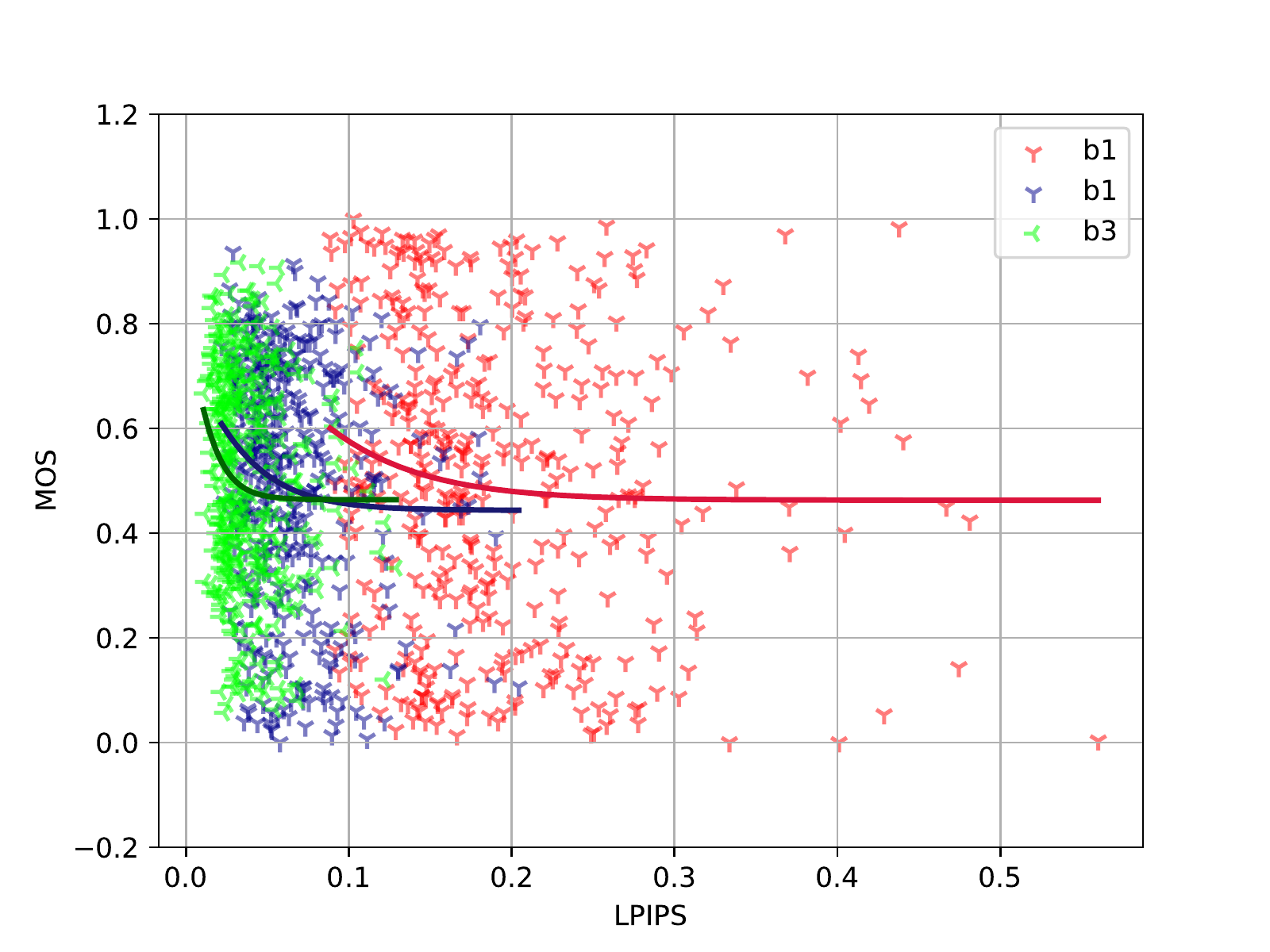}}
    \subfigure[]{\includegraphics[width = 5.13cm]{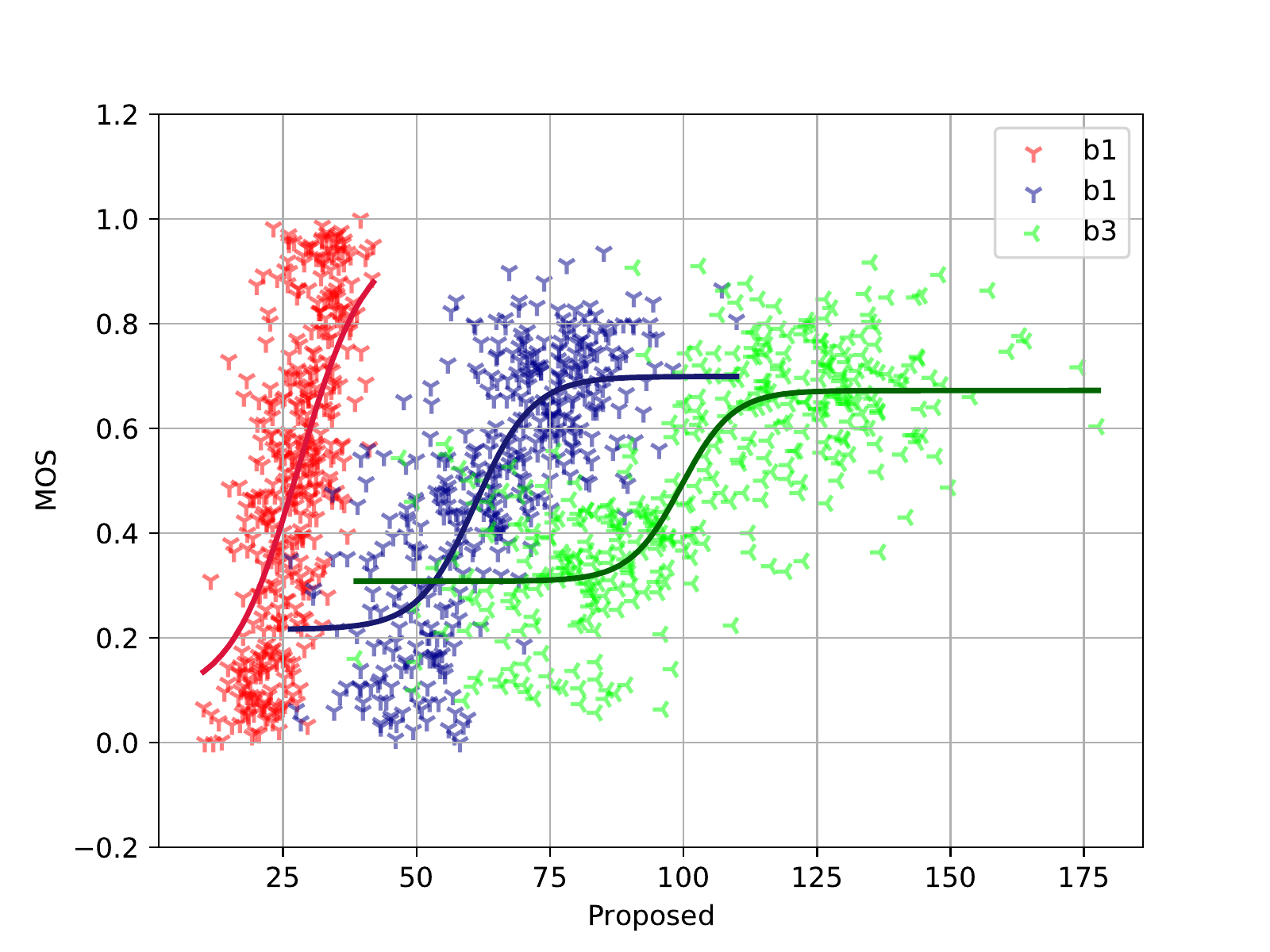}}
    \caption{Scatter plots of all compared methods on the FGIQA database \cite{fgiqa}. The red, blue, and green lines are curves fitted with a five-parameter logistic function for bit rates b1, b2, and b3. }
    \label{fig:scatter}
  \end{figure*}

\section{Experiment Validation}

\label{sec:experiment}

\begin{table*}[h]
\centering
\renewcommand\arraystretch{1.5}
\caption{Performance results of ablation experiment.}
\begin{tabular}{c|c|cccccccccc}
\toprule
 Criteria & Group & G1 & G2 & G3 & G4 & G5 & G6 & G7 & G8 & G9                       \\ \hline
\multirow{4}{*}{SRCC} & b1     & 0.9133 &0.9425 &0.6330 &0.9485 &0.9405 &0.9345 &0.9065 &0.9405 &\textbf{0.9505}                       \\
                      & b2     & 0.8460 &0.8700 &0.8088 &\textbf{0.8760} &0.8649 &0.8620 &0.8460 &0.8720 &\textbf{0.8760}                    \\
                      & b3     & 0.8331 &0.8271 &0.7731 &0.8051 &0.7871 &0.8391 &0.8491 &0.8351 &\textbf{0.8671}                       \\
                      & Average& 0.8641 &0.8799 &0.7383 &0.8765 &0.8675 &0.8785 &0.8672 &0.8825 &\textbf{0.8978}              \\ \hline
\multirow{4}{*}{KRCC} & b1     & 0.8837 &0.9041 &0.5431 &0.9147 &0.9144 &0.8907 &0.8507 &0.9007 &\textbf{0.9174}                       \\
                      & b2     & 0.7901 &0.8167 &0.7261 &\textbf{0.8267} &0.8098 &0.8034 &0.7834 &0.8167 &0.8201                       \\
                      & b3     & 0.7935 &0.7535 &0.6802 &0.7168 &0.7002 &0.7735 &0.7902 &0.7668 &\textbf{0.8035}                      \\
                      & Average& 0.8224 &0.8248 &0.6498 &0.8192 &0.7991 &0.8225 &0.8081 &0.8281 &\textbf{0.8470}              \\ \hline

\multirow{4}{*}{PLCC} & b1     & 0.9236 &0.9255 &0.7045 &0.9532 &0.9529 &0.9510 &0.9266 &0.9252 &\textbf{0.9551}                      \\
                      & b2     & 0.8902 &0.9117 &0.9049 &0.9013 &\textbf{0.9133} &0.9096 &0.9037 &0.9113 &0.9096                     \\
                      & b3     & 0.8459 &0.8705 &0.8510 &0.8619 &0.8647 &0.8713 &0.8730 &0.8711 &\textbf{0.8749}                      \\
                      & Average& 0.8865 &0.9026 &0.8201 &0.9055 &0.9103 &0.9106 &0.9011 &0.9025 &\textbf{0.9132}              \\ 
                      \bottomrule
\end{tabular}
\label{tab:ablation}
\end{table*}

\begin{table*}[h]
\centering
\renewcommand\arraystretch{1.5}
\caption{The averaged SRCC results of different values of ($\alpha$,$\beta$) on the FGIQA database. }
\begin{tabular}{c|cccccccccc}
    \toprule[0.4mm]
\diagbox [width = 3em,height = 2.5em] {$\boldsymbol{\alpha}$}{$\boldsymbol{\beta}$} & 0.1 & 0.2 & 0.3  & 0.4 & 0.5 & 0.6 & 0.7 & 0.8 & 0.9 & 1.0 \\
\hline
0.1 & 0.8825 & 0.8881 & 0.8895 & 0.8932 & 0.8939 & {\bf 0.8978} & 0.8941 & 0.8925 & 0.8919 & 0.8919  \\
0.2 & 0.8578 & 0.8825 & 0.8891 & 0.8881 & 0.8894 & 0.8894 & 0.8925 & 0.8931 & 0.8938 & 0.8938 \\
0.3 & 0.8458 & 0.8698 & 0.8825 & 0.8871 & 0.8885 & 0.8881 & 0.8901 & 0.8888 & 0.8894 & 0.8918  \\
0.4 & 0.8365 & 0.8578 & 0.8731 & 0.8825 & 0.8865 & 0.8891 & 0.8891 & 0.8881 & 0.8894 & 0.8894  \\
0.5 & 0.8284 & 0.8471 & 0.8645 & 0.8771 & 0.8825 & 0.8871 & 0.8905 & 0.8878 & 0.8891 & 0.8881 \\
0.6 & 0.8257 & 0.8458 & 0.8578 & 0.8698 & 0.8785 & 0.8825 & 0.8858 & 0.8871 & 0.8891 & 0.8885 \\
0.7 & 0.8210 & 0.8418 & 0.8525 & 0.8625 & 0.8731 & 0.8778 & 0.8825 & 0.8865 & 0.8865 & 0.8898 \\
0.8 & 0.8117 & 0.8365 & 0.8445 & 0.8578 & 0.8671 & 0.8731 & 0.8785 & 0.8825 & 0.8838 & 0.8865 \\
0.9 & 0.8070 & 0.8348 & 0.8458 & 0.8525 & 0.8625 & 0.8698 & 0.8751 & 0.8791 & 0.8825 & 0.8838 \\
1.0 & 0.8063 & 0.8284 & 0.8431 & 0.8471 & 0.8578 & 0.8645 & 0.8718 & 0.8771 & 0.8798 & 0.8825  \\
\bottomrule[0.3mm]
\end{tabular}
\label{tab:parameter}
\end{table*}

\subsection{Validated Database}
The proposed method is mainly validated on the fine-grained IQA database (FGIQA) \cite{fgiqa}, which is specially built for compressed images with fine-grained quality differences. In the FGIQA database, 100 reference images are selected and compressed into three constant target bit rates (1.96 bpp, 2.37 bpp, 2.67 bpp) by four different JPEG optimization methods, which generates 1,200 ($100 \times 3 \times 4$) compressed images in total. To distinguish subtle differences, the pair-wise comparison subjective experiments are conducted by 30 participants for every group of 4 compressed images. The final quality ranks of the fine-grained compressed images are generated by the Bradley-Terry model \cite{bt} from the pair-wise comparisons. 

\subsection{Experiment Criteria and Setup}
To test the effectiveness of the IQA methods for fine-grained compression quality assessment tasks, 3 mainstream criteria are selected for comparing the correlation between the predicted scores and ranks, which include Kendall’s Rank Order Correlation Coefficient (KRCC), Spearman Rank Correlation Coefficient (SRCC), Pearson Linear Correlation Coefficient (PLCC). Specifically, KRCC describes the similarity of the orderings, SRCC is used to measure the correlation of ranks, and PLCC denotes the linear correlation. Note that an excellent model should obtain values of SRCC, KRCC, and PLCC close to 1. 

Considering that the pair-wise comparison subjective experiments are conducted within each group of 4 compressed images, the KRCC, SRCC, and PLCC values are calculated for each group as well. Then, the average KRCC, SRCC, and PLCC values of all compressed image groups for certain bit rate are recorded as final experimental results. For example, we calculate the KRCC values of all 100 compressed image groups for b1 bit rate and record the average result as the final KRCC value for b1.

We set the weight $W_{ga}$ described in Eq. (\ref{equ:wa}) as (0.5, 0.75, 1, 5, 6). 
The weight parameters are set to increase the influence of high frequency texture because high-frequency texture usually has a greater impact on visual quality than low-frequency texture for fine-grained compressed images. $f$ and $o$ represent the serial number of frequency and orientations. Considering that the HVS is more sensitive to the $Y$ channels and the perceived difference of the other two channels is roughly the same, the weights ($W_{Y}$, $W_{Cb}$, $W_{Cr}$) for $YCbCr$ components are empirically set as (1, 0.25, 0.25).
Through experiment, we obtain the optimal set of $\alpha$ and $\beta$ as (0.1, 0.6) for the FGIQA database, which is further discussed in Section \ref{sec:parameter}.

Additionally, we employ the similar procedures described in \cite{statistic-test} to fit the predicted scores to real quality values with a five-parameter logistic function which can be derived as:
\begin{equation}
V =\tau_{1}\left(\frac{1}{2}-\frac{1}{1+e^{\tau_{2}\left(q-\tau_{3}\right)}}\right)+\tau_{4} s+\tau_{5}
\end{equation}
where $V$ and $q$ represent the fitted scores and predicted scores, $\left\{\tau_{i} \mid i=1,2, \ldots, 5\right\}$ are the corresponding parameters.

\subsection{Experiment Competitors}
To compare the effectiveness of different IQA methods, we choose several state-of-the-art FR-IQA methods for comparison, which can be categorized into two types:
\begin{itemize}
    \item Hand-crafted-based methods: PSNR, SSIM \cite{ssim}, VSI \cite{vsi}, MS-SSIM \cite{ms-ssim}, IW-SSIM \cite{iw-ssim}, FSIM \cite{fsim}, MAD \cite{mad}, GMSD \cite{gsmd}, and IFC \cite{ifc}.
\end{itemize}
\begin{itemize}
    \item CNN-based methods: PieAPP \cite{pieapp} and LPIPS \cite{lpips}. Specifically, these CNN-based methods are tested with default parameters.
\end{itemize}


\subsection{Experimental Performance}
The experimental results of the proposed methods and other selected state-of-the-art FR-IQA models are listed in Table \ref{tab:fgiqa}, where b1, b2, b3 represent the three target bit rates and the top 2 results are marked in bold for each column. According to Table \ref{tab:fgiqa}, several observations can be made. First, the proposed method achieves top 2 places in KRCC, SRCC, PLCC among all three target bit rates and obtains the best performance on average, which indicates that our method is more effective for fine-grained compressed image quality assessment. PSNR and SSIM are the two most widely used models for compression systems, however, they fail to accurately predict the quality levels of fine-grained compressed images. The reason may be that PSNR and SSIM are developed and validated on traditional IQA databases where the distortions are relatively simple and the images are usually distorted with coarse-grained levels. Then, it can be seen that although some compared traditional FR-IQA methods achieve good performance on certain bit rate, they may be less effective for other bit rate, meaning that they could be less stable when the bit rate keeps changing. What's more, the CNN-based methods perform at an intermediate level. This is because that the models are trained on databases with coarse-grained quality levels, which results in their being less sensitive to fine-grained compressed images' artifacts.

The scatter plots of the proposed method and compared FR-IQA models are illustrated in Fig. \ref{fig:scatter}, where the red, blue, and green lines are curves fitted with a five-parameter logistic function for bit rates b1, b2, and b3.  We can observe that the scatter points are more clustered to the proposed methods' fitted curves for all three bit rates of the FGIQA database, which strongly validates the excellent prediction ability of the proposed method. What's more, statistical significance tests are conducted to further verify the performance of the proposed method. We use t-test as described in \cite{statistic-test} to compare the difference between the predicted quality scores with the subjective ratings. The null hypothesis of the t-test is set as the residuals of two quality metrics derived from the same distribution and they are statistically indistinguishable with a 95\% confidence. The statistical significance tests results are shown in Fig. \ref{fig:heatmap}, from which we can see the proposed method is significantly superior to most compared methods.

\subsection{Experiment Results on Other Compression Databases}
To further test the generalization ability of the proposed method on coarse-grained compressed images, we select some popular IQA databases for validation, which include LIVE \cite{live}, TID2013 \cite{tid2013}, and CSIQ \cite{csiq}. The well-known LIVE database contains 9 reference images and 779 distorted images with five distortion types. The TID2013 database contains 25 reference images and utilizes 24 types of distortions while the CSIQ database consists of 30 reference images and employs six types of distortions. All the IQA databases mentioned above include distortion types of JPEG and JPEG2000, and we only test our method on the compression distorted images. The parameters are set the same as discussed previously.
The final results are shown in Table \ref{tab:others}, where the methods with the best performance are marked in bold. Some IQA methods that are competitive on the FGIQA database are also tested for comparison. Although the proposed method achieves first place only on the LIVE compression database, it still outperforms other IQA methods on average, which proves the effectiveness and stability of the proposed method.

\subsection{Ablation Experiment}
\label{sec:parameter}
To analyze the contributions of different groups of features and different scales of texture, we conduct the ablation experiment in this section. Specifically, ablation groups are defined as:
\begin{itemize}
    \item G1: Without gradient-based features;
    \item G2: Without region selection of gradient-based features;
    \item G3: Without texture-based features;
    \item G4: Without texture-based frequency scale 1 features;
    \item G5: Without texture-based frequency scale 2 features;
    \item G6: Without texture-based frequency scale 3 features;
    \item G7: Without texture-based frequency scale 4 features;
    \item G8: Without texture-based frequency scale 5 features;
    \item G9: With all features.
\end{itemize}

The results of the ablation experiment are shown in Table \ref{tab:ablation}, where the best performance is marked in bold for each row. It can be clearly observed that G9 outperforms all other groups, which indicates that gradient-based features and texture-based features all contribute to the final results. By comparing the performances of G1 and G3, we can find that G1 performs better than G2, meaning that texture-based features make more contributions. G9's performance is superior to G2, meaning that the region selection of gradient-based features can help improve the performance. Additionally, it can be seen that G5 and G7 achieve relatively lower performance among G4-G8, which indicates that texture-based features from frequency scales 2 and 4 make more contributions than other frequency scales.

\subsection{Parameter Sensitivity}
Noticing that the proposed method includes two main parameters ($\alpha$, $\beta$) in Eq. (\ref{equ:parameter}), we decide to conduct the parameter sensitivity experiment in this section. These two parameters are employed to determine the significance of gradient-based features and texture-based features. Here, we change the values of ($\alpha$, $\beta$) to test the performance of the proposed method and obtain the optimal values of ($\alpha$, $\beta$). More specifically, we vary the parameters ($\alpha$, $\beta$) both from 0.1 to 1 with a step of 0.1 and the parameters are changed at the same time. The experimental results of averaged SRCC are shown in Table \ref{tab:parameter}, from which we can see that the proposed method gains good stability over the range of test parameters and the optimal values for ($\alpha$, $\beta$) are (0.1,0.6).


\section{Conclusion}
\label{sec:conclusion}
In this paper, we discuss the quality assessment problems of fine-grained compressed images. Fine-grained compressed images indicate the images that are compressed with close or the same bit rate. These images' distortion differences are quite subtle, thus bringing a huge challenge for the corresponding quality assessment. Most state-of-the-art full-reference metrics are developed on the image quality assessment databases with coarse-grained compression levels, therefore, they may be not effective for dealing with fine-grained compression quality assessment.

Normally speaking, people usually perceive the compressed images in both coarse and fine manners. People judge the quality both through the obvious distortions and indistinct details of compressed images. Thus, we propose a full-reference metric to especially assess the quality levels of fine-grained compressed images. We extract coarse and fine features from two aspects: image gradient and texture.  The gradient-based features are sensitive to obvious distortions while the texture-based features can identify tiny quality differences. The gradient-based features are extracted by calculating the statistical parameters of the gradient similarity maps in regions selected by frequency information. The texture-based features are extracted by computing the statistical parameters of the Log-Gabor similarity maps with different frequency levels and orientations. Finally, the quality score is obtained by fusing the features mentioned above. The experimental results show that our method outperforms all compared methods on the FGIQA database and achieves competitive performance on other coarse-grained compression quality assessment databases. The ablation experiment and statistical test further confirm the effectiveness of the proposed method. We hope our method can help promote the development of compression systems.

\bibliographystyle{IEEEtran}
\bibliography{references}

\end{document}